\documentclass[preprint,authoryear]{elsarticle}
\makeatletter
\def\ps@pprintTitle{%
  \let\@oddhead\@empty\let\@evenhead\@empty
  \def\@oddfoot{\reset@font\hfil\thepage\hfil}%
  \let\@evenfoot\@oddfoot}
\makeatother

\usepackage{lineno}
\usepackage{pifont}
\usepackage{graphicx}
\usepackage{multirow}
\usepackage{amsmath,amssymb,amsfonts}
\usepackage{amsthm}
\usepackage{xcolor}
\usepackage{textcomp}
\usepackage{manyfoot}
\usepackage{booktabs}
\usepackage[linesnumbered,ruled,vlined]{algorithm2e}
\usepackage{listings}
\usepackage[utf8]{inputenc}
\usepackage[T1]{fontenc}
\usepackage{subcaption}
\usepackage{xurl}
\usepackage{hyperref}

\theoremstyle{definition}

\newtheorem{definition}{Definition}

\begin{document}

\begin{frontmatter}

\title{Shaping the Evolutionary Dynamics of Robot Morphology via Adaptive Control Learning}

\author[a]{Junru Song\fnref{fn2}}
\author[b]{Yang Yang\fnref{fn2}}
\author[c]{Yaqing Xu}
\author[a,d]{Ying Wen}
\author[b]{Wei Peng}
\author[b]{Guozhen Li}
\author[b]{Wei'en Zhou\fnref{fn1}}
\ead{weienzhou@outlook.com}
\author[b]{Wen Yao\fnref{fn1}}
\ead{wendy0782@126.com}

\address[a]{Shanghai Jiao Tong University, Shanghai 200240, China}
\address[b]{Intelligent Game and Decision Laboratory, Beijing 100048, China}
\address[c]{Renmin University of China, Beijing 100872, China}
\address[d]{Shanghai Innovation Institute, Shanghai 200232, China}

\fntext[fn1]{Corresponding author.}
\fntext[fn2]{Co-first authors with equal contribution.}

\begin{abstract}
Robot co-design via bi-level optimization couples within-lifetime controller learning for fitness evaluation with cross-generational morphological evolution. Prior work has established that well-adapted morphology facilitates faster control learning, a property termed \emph{morphological intelligence}. Yet how control learning reciprocally shapes morphological evolution remains unexplored. This paper examines both directions for a holistic account of brain-body interplay. We first show that morphological contributions to control learning decouple into two orthogonal dimensions. We formalize the convergence speed as \emph{morphological intelligence} and identify the performance ceiling as a complementary quantity termed \emph{true potential}. A concise functional relation is then established to jointly characterize both quantities from individual learning curves, which, when aggregated at the population level, capture evolutionary profiles. Through extensive experiments on simulated voxel-based soft robots, we reveal that premature fitness evaluation systematically underestimates true potential and biases selection towards fast learners. This restricts design space exploration, compromising both optimization efficiency and morphological diversity. Notably, the widely recognized \emph{morphological Baldwin effect} emerges as an artifact of this bias rather than a general evolutionary tendency. We therefore propose AdaControl, which monitors disproportionate selection for morphological intelligence during evolution and allocates minimally sufficient control learning for unbiased fitness evaluation. With AdaControl, a simple genetic algorithm rivals state-of-the-art generative-model-based co-design methods in discovering diverse high-performing designs while cutting computation by up to 80\% versus exhaustive control. More broadly, we reveal how learning and evolution interact across timescales to shape embodied intelligence, shedding light on the whole picture of brain-body co-evolution.
\end{abstract}


\end{frontmatter}

\section{Introduction}
\label{sec:intro}

The emergence of intelligence and adaptive behavior is fundamentally rooted in the interplay between brain, body, and environment \citep{buason2005brains}. Living organisms exploit their morphology and its interaction with the environment to substantially reduce the cognitive demands of behavioral control \citep{li2016fish, ghazi2019morphological}. Drawing on this insight, robotics research has confirmed that well-adapted robot morphology can similarly alleviate the computational demands of control, a principle formalized as \emph{morphological intelligence} \citep{ghazi2019morphological}.

Inspired by such brain-body synergy observed in nature \citep{pfeifer2006body,pfeifer2014cognition}, robot co-design seeks to jointly optimize morphology and control, typically formulated as a bi-level problem coupling two processes on distinct timescales. The inner loop optimizes a dedicated sensorimotor controller for each candidate morphology and evaluates its task performance. The outer loop employs evolutionary algorithms (EAs) to maintain and refine a population of morphological designs based on evaluated fitness. Within this framework, a landmark finding is the \emph{morphological Baldwin effect} \citep{gupta2021embodied}, where morphological intelligence was observed to increase monotonically throughout evolution, widely cited as evidence that evolution inherently favors morphologies that learn faster. However, the interaction between evolution and learning is not one-directional, as the latter provides essential fitness signals that govern which morphologies survive and proliferate \citep{eiben2020if, mertan2025evolutionary}. Nevertheless, most co-design studies treat control configurations as fixed, subjective design choices \citep{bhatia2021evolution,hu2022modular,hu2023glso,song2024morphvae,liu2025cdmeo,wang2026gnn,rossi2026distributed}, overlooking how they could shape morphological evolution across generations. \citet{goff2021challenges} offered preliminary evidence that simpler controllers yield higher morphological diversity, but this study was confined to rigid robots, flat-terrain locomotion, and controllers with only a few dozen parameters, leaving its relevance to more general settings unclear.

In this work, we address this gap by jointly examining both timescales. A central insight we provide is that the control learning curve of an individual robot reflects not only the quality of its controller but also fundamental properties of the underlying morphology. We formalize this observation through a concise functional relation that decouples morphological contributions to control learning into two orthogonal quantities. Specifically, \emph{morphological intelligence} captures the convergence speed of control learning afforded by a morphology, while \emph{true potential} represents its performance ceiling. These quantities can be estimated from observed learning curves through simple non-linear regression and, when aggregated at the population level, yield quantitative indicators of evolutionary behaviors.

We base our experiments on simulated voxel-based soft robots (VSRs) \citep{hiller2011automatic,bhatia2021evolution}. VSRs are composed of elastic cubic blocks interconnected in a grid-like layout and achieve motion through volumetric actuation. The compliance of soft materials gives rise to rich inter-voxel and morphology-environment interactions. This produces sophisticated evolutionary landscapes that well represent real-world co-design challenges \citep{mertan2025evolutionary}. Using the genetic algorithm (GA) \citep{michalewicz2013genetic} and proximal policy optimization (PPO) \citep{schulman2017proximal}, canonical choices for morphological evolution and control learning in the literature \citep{bhatia2021evolution,song2024morphvae,zhao2025cross}, we conduct extensive experiments across a spectrum of control complexities. We report a key finding: prematurely terminated control learning systematically underestimates true potential of candidate morphologies and biases natural selection towards robots that learn faster in their early lifetime at the expense of long-term performance. This in turn hampers design space exploration and compromises both optimization efficiency and morphological diversity of robot co-design. Notably, the morphological Baldwin effect reported in \citet{gupta2021embodied} emerges as a special case of this bias rather than a general evolutionary phenomenon.

These findings motivate AdaControl, an adaptive algorithm that schedules control learning based on observed evolutionary behavior. In each generation, AdaControl begins with a minimal learning budget and performs tentative natural selection. Leveraging the proposed quantification of morphological intelligence, it monitors whether survivors are disproportionately biased towards fast learners and progressively extends the population's learning cycle until such disparity diminishes. This largely ensures unbiased fitness evaluation at minimal computational cost, guarding against both truncated learning and indiscriminate exhaustive training. We demonstrate that, with AdaControl, a simple genetic algorithm matches the optimization efficiency of exhaustive control with up to 80\% less computation and rivals state-of-the-art co-design methods built on deep generative models, while uncovering a more diverse repertoire of high-performing morphologies.

The contributions of this work are summarized as follows:
\begin{itemize}

    \item We introduce a novel data-driven perspective that extracts the intrinsic learning profile of morphologies from control learning curves, providing quantitative tools for analyzing evolutionary dynamics at the population level.

    \item We reinterpret the morphological Baldwin effect, widely regarded as a general evolutionary principle, as an artifact of prematurely terminated control learning, identifying the configuration of control as a critical yet overlooked driver of evolutionary outcomes.

    \item Through AdaControl, we demonstrate that principled fitness evaluation is a more fundamental determinant of robot co-design performance than the choice of search strategy, as validated extensively on simulated VSRs. Our findings pave the way for more scalable robot co-design and offer new insights into brain-body co-evolution in learning-based robotic systems.

\end{itemize}

The rest of this paper is organized as follows. Section \ref{sec:pre} introduces voxel-based soft robots, the co-design algorithm, and morphological intelligence. Section \ref{sec:theory} presents the mathematical framework to formalize the learning profile of morphologies, with motivating experiments. Section \ref{sec:method} details AdaControl. Section \ref{sec:exp} reports experimental results and analyses. Section \ref{sec:conclusion} concludes the paper.

\section{Preliminaries}
\label{sec:pre}
\subsection{Voxel-Based Soft Robots}

\begin{figure*}
\centering
\includegraphics[width=\textwidth]{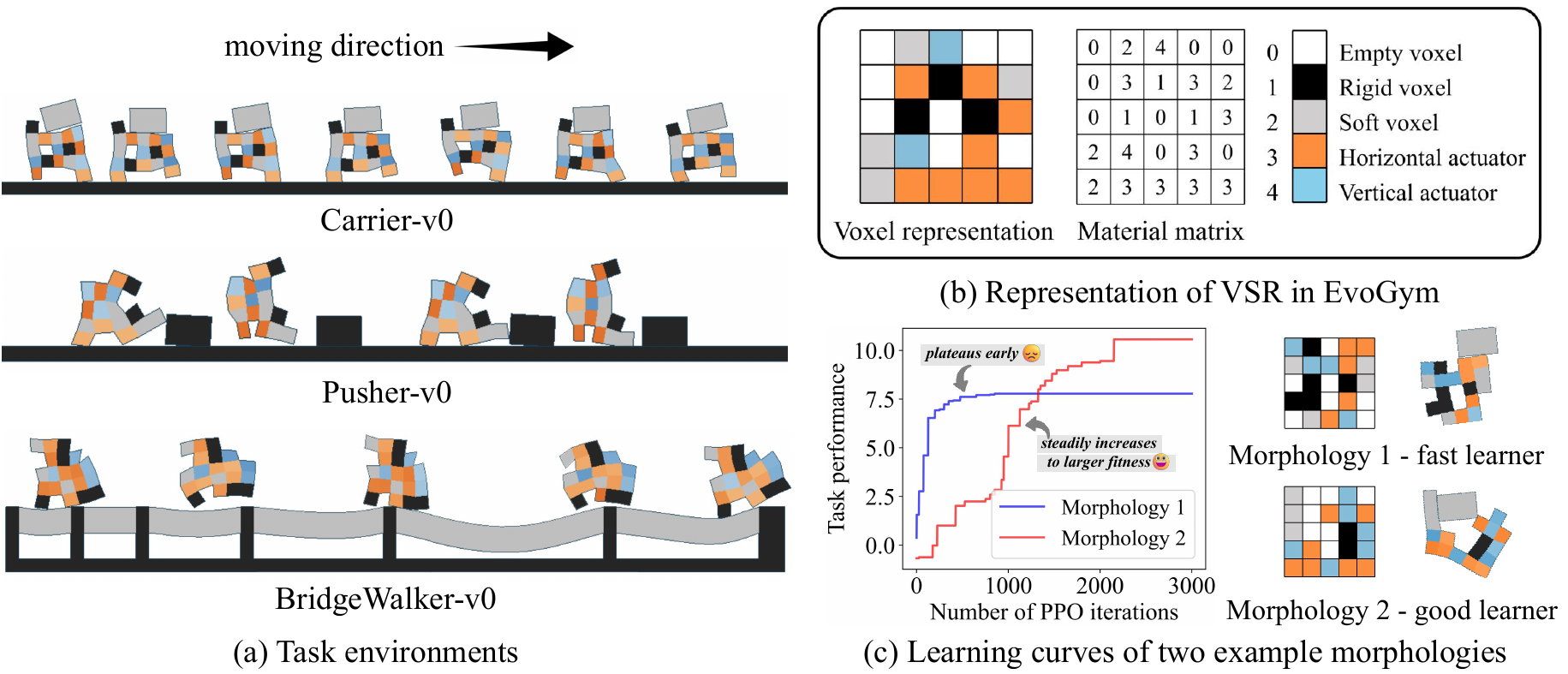}
\caption{{(a) Task environments used for benchmarking. (b) Two-dimensional VSR representation in EvoGym \citep{bhatia2021evolution}. (c) Two representative morphologies from preliminary experiments, illustrating the distinction between fast and good learners. } \label{Fig.1}}
\end{figure*}

Voxel-based soft robots (VSRs) consist of elastic cubic blocks, or \emph{voxels}, interconnected in a grid-like layout. Unlike rigid robots with articulated limbs, VSRs exploit the compliance of soft materials to achieve far greater degrees of freedom, with motion driven by volumetric actuation of designated \emph{actuator} voxels (see Fig. \ref{Fig.1}(a)). This compliance makes VSRs particularly suited to unstructured environments requiring adaptability. However, it also gives rise to complex interactions among voxels and between the robot and its environment, producing a rich yet underexplored interplay between morphology and control.

Numerous simulation platforms have been developed for VSRs \citep{huang2020dynamic,bhatia2021evolution,huang2021plasticinelab,dubied2022sim,medvet20202d,li2024generating,shen2026evogymcm}. We adopt Evolution Gym (EvoGym) \citep{bhatia2021evolution} for its lightweight physics engine, accessible Python interface, and diverse task suite. As illustrated in Fig. \ref{Fig.1}(b), EvoGym supports five material types to span an expressive design space: empty (0), rigid (1), soft (2), horizontal actuator (3), and vertical actuator (4). The underlying physics is modeled through a mass-spring system with cross-braced springs and penalty-based frictional contact \citep{bhatia2021evolution}.

We evaluate on three EvoGym task environments: Carrier-v0, Pusher-v0, and BridgeWalker-v0 (Fig. \ref{Fig.1}(a)), hereafter referred to as Carrier, Pusher, and BridgeWalker for brevity. The first two involve transporting a rectangular box through different strategies, while the third requires locomotion across a deformable terrain. Together they span manipulation and locomotion, two canonical application domains, and serve as established benchmarks in the co-design literature \citep{bhatia2021evolution,wang2023preco,songlaser,liu2025cdmeo,fang2025robomore}.

Our work relies on simulated VSRs due to the prohibitive cost of physical fabrication. Nevertheless, recent advances in soft robot manufacturing, including pneumatic polymer chambers \citep{kriegman2020bscalable,legrand2023reconfigurable} and biologically based self-replicating systems \citep{kriegman2020ascalable,kriegman2021kinematic}, are narrowing the sim-to-real gap. We expect the insights developed here to inform future studies on the evolutionary dynamics of physical soft robots.

\subsection{Robot Co-design}
\begin{figure}[h]
\centering
\includegraphics[width=0.65\textwidth]{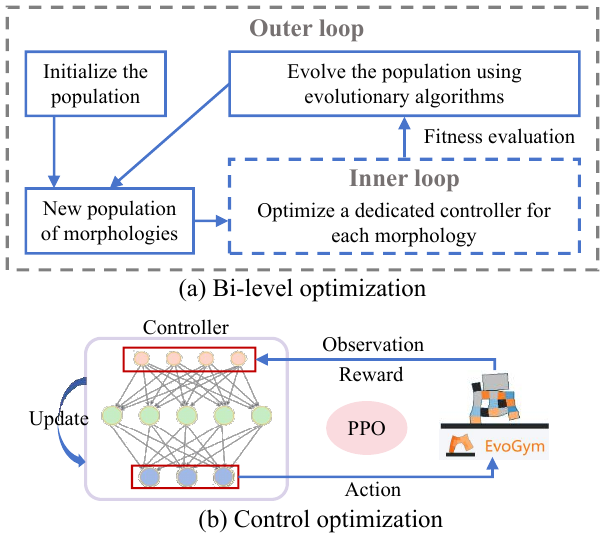}
\caption{{Robot co-design. } \label{Fig.2}}
\end{figure}

Robot co-design jointly optimizes morphological designs and sensorimotor controllers to achieve intelligent behaviors. This is typically formulated as a bi-level optimization problem (Fig. \ref{Fig.2}(a)):
\begin{equation}
    x^*=\arg\max_{x\in\mathcal{X}}f(x,c^*),
\end{equation}
\begin{equation}
    \text{s.t. }c^*=\arg\max_{c\in \mathcal{C}}f(x,c),
\end{equation}
where $\mathcal{X}$ and $\mathcal{C}$ denote the morphology and controller spaces, respectively, and $f(x,c)$ evaluates the task performance of morphology $x$ under controller $c$. The inner and outer loops are detailed below.

\subsubsection{Inner loop}

The inner loop optimizes a dedicated controller for each candidate morphology (Fig. \ref{Fig.2}(b)). When controllers are parameterized as deep neural networks, reinforcement learning (RL) is the predominant optimization method. RL proceeds by alternating between environment sampling and policy updates, with each such cycle termed an \emph{iteration}. The resulting task performance, measured as the cumulative reward over a complete episode, serves as the fitness of the morphology. We parameterize each controller as a multi-layer perceptron (MLP) that maps environmental observations to actuation signals, one per actuator voxel. In EvoGym, each signal drives an expansion or contraction of the corresponding actuator relative to its rest volume, transitioning the environment to the next state and closing the perception-action loop.

\subsubsection{Outer loop}
The outer loop evolves morphological designs using evolutionary algorithms such as the genetic algorithm (GA) \citep{michalewicz2013genetic}, Bayesian optimization \citep{pelikan2005bayesian}, and CPPN-NEAT \citep{stanley2007compositional}. A population of morphologies is maintained and iteratively refined through stochastic variation, guided by fitness scores from the inner loop. We adopt the GA variant adapted for VSRs in \citet{bhatia2021evolution}. In each generation, morphologies are evaluated and ranked by fitness. Top-ranking individuals survive and undergo random voxel mutations to produce the next generation. This cycle repeats until a pre-specified evaluation budget is exhausted.

Note that PPO \citep{schulman2017proximal} and GA \citep{michalewicz2013genetic} are both canonical algorithms widely adopted in robot co-design, intentionally selected here to ensure the generality of our findings. The reader is referred to \citet{bhatia2021evolution} for implementation details.

A major bottleneck in co-design is the expensive per-morphology controller optimization in the inner loop. Existing remedies include action inheritance \citep{liu2023rapidly} and policy transfer \citep{liu2024meta, chen2024mirage}. A more radical alternative is Lamarckian inheritance, where a shared universal controller \citep{gupta2022metamorph, strgar2025accelerated} is updated and passed across generations. However, such strategies have been found prone to premature convergence, as the inherited controller favors morphologies with first-mover advantage in control and fails to generalize to novel designs \citep{mertan2024investigating, mertan2025evolutionary}. These findings highlight that the configuration of control learning can profoundly shape evolutionary outcomes, and that dedicated per-morphology control optimization remains the most reliable paradigm in co-design, based on which our work is conducted.

\subsection{Morphological Intelligence}
\label{sec:intel}
The concept of \emph{morphological computation}, introduced in the early 2000s \citep{maass2002real,pfeifer2006body}, originally characterized how physical body-environment interactions can perform computations that would otherwise burden the controller. However, the term gradually became narrowly associated with physical reservoir computing \citep{hauser2011towards, fuchslin2013morphological, muller2017morphological}. To address this limitation, \citet{ghazi2019morphological} proposed \emph{morphological intelligence} as a broader framework, defined as ``the reduction of computational cost for the brain (or controller) resulting from the exploitation of the morphology and its interaction with the environment.'' \citet{woodward2018morphological} instantiated this concept as the reduction of slipping events on uneven terrains through passive mechanics. In the context of learning-based control, \citet{gupta2021embodied} assessed morphological intelligence through the speed and performance of reinforcement learning, and observed an ever-increasing trend throughout evolution termed the \emph{morphological Baldwin effect}.

Quantifying morphological intelligence remains an open challenge. \citet{ghazi2019morphological} introduced a causal model of sensorimotor loops and employed information-theoretic methods to measure morphological contributions to control. Other approaches \citep{polani2011informational,ruckert2013stochastic} frame the problem as an optimization task, examining how much control complexity can be reduced while preserving intelligent behavior. These methods, however, are generally confined to simplified dynamic systems or require computationally involved analysis, limiting their applicability to learning-based soft robotic systems. Inspired by the perspective of \citet{gupta2021embodied}, we propose to quantify morphological intelligence directly from control learning curves via non-linear regression and introduce a complementary quantity, \emph{true potential}, capturing the performance ceiling of a morphology. Together, these provide a complete characterization of a morphology's intrinsic learning profile that can be aggregated at the population level to track evolutionary trends, as detailed in Section \ref{sec:theory}.

\section{Morphological Intelligence and True Potential}
\label{sec:theory}
\subsection{Qualitative Analysis}
\label{sec:qualitative}
As discussed in Section \ref{sec:intel}, \citet{gupta2021embodied} assessed morphological intelligence through the speed and performance of reinforcement learning. However, the RL learning curves reported in \citet{gupta2021embodied} exhibit clear rising trends at termination, suggesting that control learning was cut short before convergence. This means that while task performance achieved within fixed iterations does reflect morphological quality, it captures only part of the picture. The maximal performance attainable after full convergence, which we term the \emph{true potential} of a morphology, is an equally important yet distinct dimension of morphological quality.

Crucially, fast learning does not imply high true potential. To verify this, we conduct a preliminary experiment on Carrier following the GA implementation of \citet{bhatia2021evolution}, but extend PPO training to 3000 iterations to ensure convergence. Fig. \ref{Fig.1}(c) shows the learning curves of two representative morphologies (quantitative results in Section \ref{sec:quant}). Morphology 1 is a fast learner that reaches decent performance early but plateaus at a modest level. Morphology 2 learns more slowly but ultimately achieves substantially higher performance. Notably, at 1000 iterations, a commonly adopted setting in prior work, Morphology 1 outperforms Morphology 2, rendering the latter's true potential invisible to the selection process.

\subsection{Definitions}
Building on the above observations, we formally define the two core concepts of this work, in the context of learning-based robotic systems.

\begin{definition}[True potential]
The \textbf{true potential} of a robot morphology is the upper limit of task performance attainable after control learning fully converges.
\end{definition}

\begin{definition}[Morphological intelligence, MI]
The \textbf{morphological intelligence (MI)} of a robot morphology is the convergence speed of its control learning process towards its true potential.
\end{definition}

In the RL setting, task performance corresponds to cumulative episodic reward, and morphological intelligence to the convergence rate of the learning curve. Our definitions refine the perspective of \citet{gupta2021embodied}, who assessed morphological intelligence as ``the speed and performance of reinforcement learning.'' We disambiguate \emph{performance}, which in \citet{gupta2021embodied} was evaluated at fixed and arbitrary iteration counts, into the maximal performance after convergence, termed true potential. This separation yields two orthogonal dimensions of morphological contribution to intelligent behavior, enabling more systematic analysis of their respective roles and interactions.

\subsection{Quantification}
We now formalize the above concepts mathematically. Let $\alpha$, $A$, and $C$ denote morphological intelligence, true potential, and control complexity, respectively. Control complexity can in principle be modulated through either network architecture or training duration. Here we adopt the latter, defining $C$ as the number of RL iterations, which allows different complexities to be compared using the same network as it is progressively trained. We model the dependence of task performance on morphology and control with a hyperbolic tangent:
\begin{equation}
\label{eq:core}
    f(x,C)=A \cdot \text{tanh}\left(  \frac{C}{C^*}\cdot \alpha\right)+\epsilon,
\end{equation}

\noindent where $x$ denotes a robot morphology and $C^*$ the number of iterations required for control learning to fully saturate. Normalizing by $C^*$ makes $\alpha$ a \emph{dimensionless} convergence rate, while $A$ shares the dimension of task performance and depends on the reward function. The noise term $\epsilon$ accounts for stochasticity in both controller optimization and environmental dynamics. The hyperbolic tangent is chosen for its simplicity and its ability to approximate the saturating nature of learning curves observed in our experiments. Alternative functional forms may be considered for scenarios with irregular convergence behavior. The parameters of interest, $\alpha$ and $A$, are estimated via non-linear regression as follows.

Both $\alpha$ and $A$ are treated as trainable parameters and initialized randomly. We minimize the mean squared error between Eq. (\ref{eq:core}) and the observed learning curve via gradient descent. For a specific morphology $x$,

\begin{equation}
\begin{aligned}
\nonumber
    A^*,\alpha^*&\leftarrow\arg\min_{A,\alpha}\frac{1}{|\mathcal{S}|}\sum_{C_i\in\mathcal{S}}(f_{i}-f(x,C_i))^2\\
    &\equiv \arg\min_{A,\alpha}\frac{1}{|\mathcal{S}|}\sum_{C_i\in\mathcal{S}}\left(f_{i}-A\cdot \text{tanh}\left(  \frac{C_i}{C^*}\cdot\alpha  \right) \right)^2, \\
    &~~\text{s.t. }\alpha>0,
\end{aligned}
\end{equation}

\noindent where $\mathcal{S}$ is the set of evaluation points along the learning curve, with $C_i$ and $f_i$ the iteration count and observed performance at the $i$-th point. To enforce $\alpha>0$, we reparameterize as $\alpha=\exp(\tilde{\alpha})$ with $\tilde{\alpha}\in\mathbb{R}$. $A$ is unconstrained, as task performance need not be positive. For notational simplicity, the estimates $A^*$ and $\alpha^*$ are hereafter written as $A$ and $\alpha$. With only two parameters, the estimation procedure incurs negligible computational overhead.

\subsection{Quantitative Analysis}
\label{sec:quant}
\begin{figure}[h]
\centering
\includegraphics[width=0.55\textwidth]{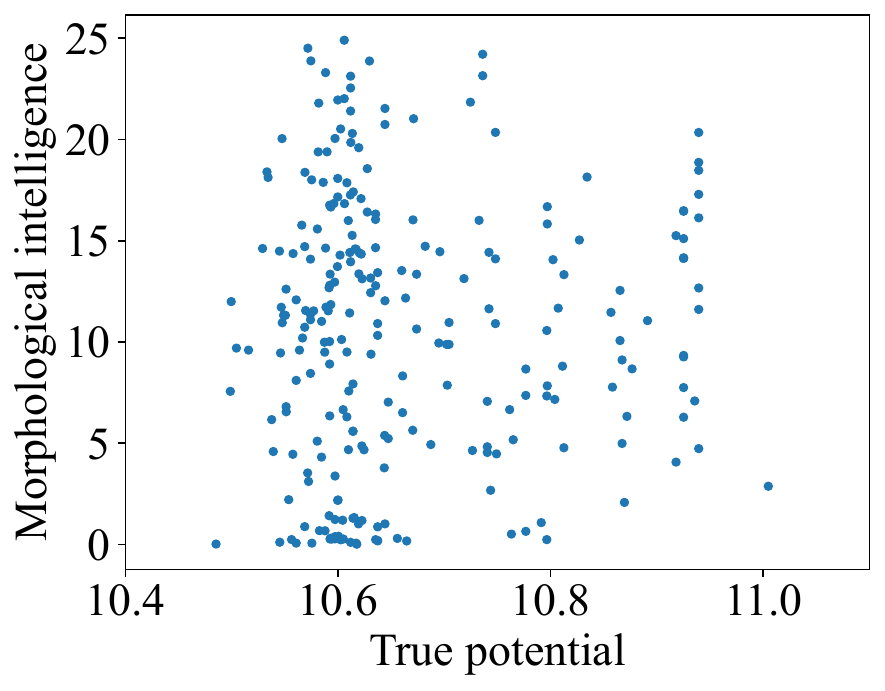}
\caption{Morphological intelligence versus true potential for 250 high-performing morphologies (top 5\% of 1,000 evaluated solutions $\times$ five independent trials). No significant correlation is observed (Spearman's $\rho=0.05$, $p=0.23$), though the variance of morphological intelligence narrows at higher true potential. \label{fig:correlation}}
\end{figure}

Using the preliminary experiment on Carrier described in Section \ref{sec:qualitative}, we estimate the morphological intelligence and true potential of all high-performing morphologies. As shown in Fig. \ref{fig:correlation}, the two quantities exhibit no significant correlation, with a Spearman's rank correlation coefficient of 0.05 ($p$-value$=0.23$). This confirms that morphological intelligence and true potential are nearly \emph{orthogonal} dimensions of morphological quality. A natural consequence is that prematurely terminated control learning would implicitly favor fast learners at the expense of morphologies with high true potential, biasing evolutionary outcomes.

\section{AdaControl}
\label{sec:method}

\begin{figure*}[h]
\centering
\includegraphics[width=0.95\textwidth]{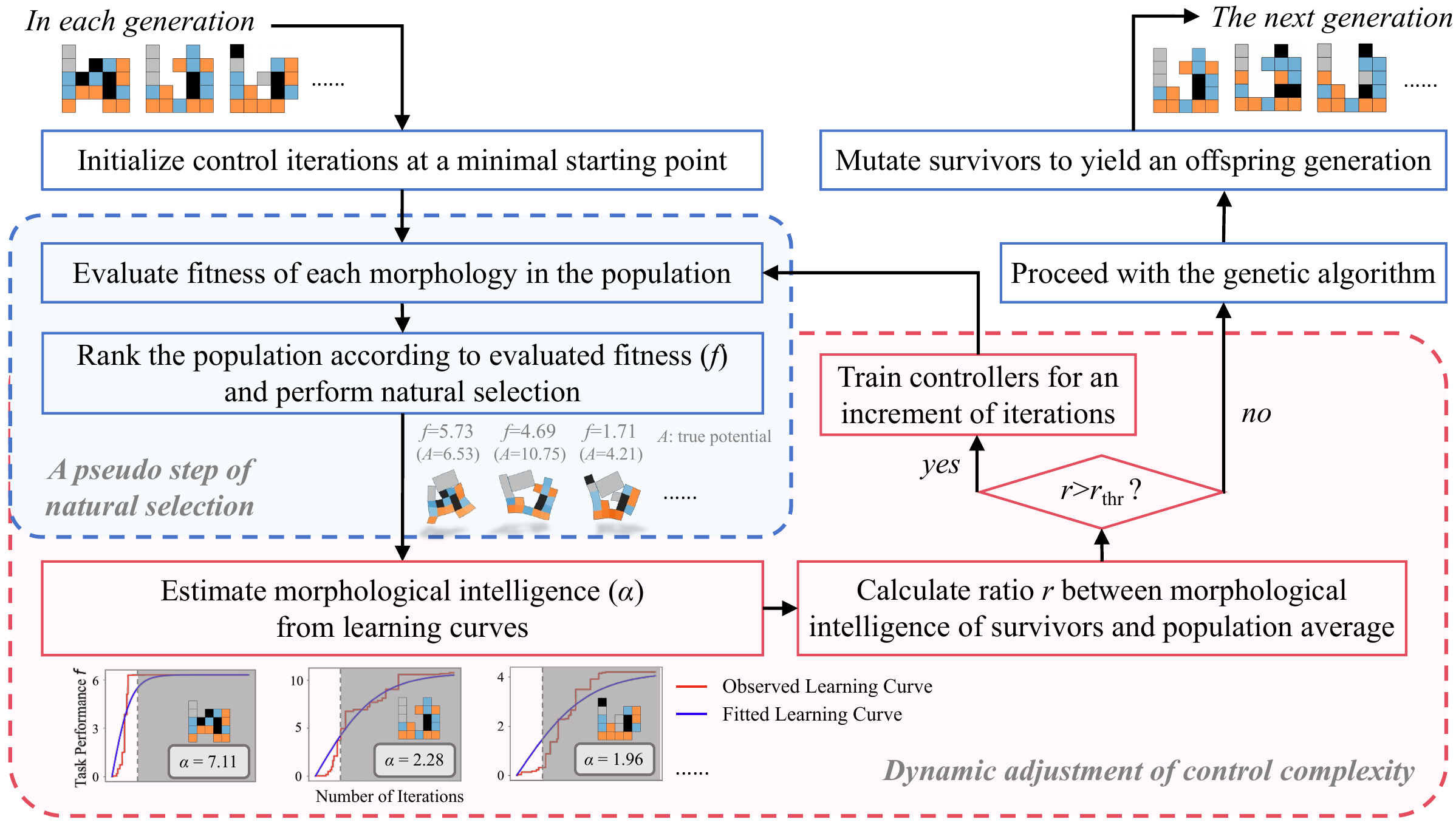}
\caption{Overview of robot co-design with AdaControl. In each generation, the population undergoes progressive control learning with iteratively increased iterations until the morphological intelligence of survivors is sufficiently close to the population average. In each \emph{pseudo step} of natural selection, only the observed segments of learning curves (unshaded areas) are used for estimating morphological intelligence. \label{Fig.4}}
\end{figure*}

In Section \ref{sec:exp}, we validate through extensive experiments that prematurely terminated control learning indeed biases natural selection towards fast learners, narrowing the evolutionary search to a restricted subspace and consequently compromising both optimization efficiency and morphological diversity. While simply prolonging control learning can mitigate this bias, it is a brute-force solution that quickly becomes infeasible under limited computational budgets. We instead propose AdaControl, an adaptive algorithm that monitors evolutionary dynamics and calibrates control complexity accordingly.

AdaControl builds on the quantification framework of Section \ref{sec:theory}. In each generation, control learning begins at a minimal number of iterations. A \emph{pseudo step} of natural selection is then performed based on currently evaluated fitness, tentatively identifying survivors. Rather than immediately proceeding to mutation, AdaControl pauses to assess potential bias. The morphological intelligence of the entire population is estimated, and the ratio $r$ of survivor-average to population-average morphological intelligence is computed. An $r$ substantially exceeding 1 signals that fast learners are disproportionately favored due to insufficient control learning. In this case, all controllers resume training for an additional increment of iterations, and the pseudo selection is repeated. This loop continues until $r$ falls below a pre-specified threshold $r_{\text{thr}}$, at which point natural selection is deemed unbiased and evolution proceeds normally. Each pseudo step incurs only negligible overhead for morphological intelligence estimation, yet ensures that control learning is neither prematurely curtailed nor wastefully prolonged. Robot co-design with AdaControl is illustrated in Fig. \ref{Fig.4} and outlined in Algorithm \ref{alg:adacontrol}.

\begin{algorithm}[t!]
\small
\caption{Robot Co-design with AdaControl\label{alg:adacontrol}}
\KwIn{Evaluation budget $M$, population size $N$, survival rate $s$, MI ratio threshold $r_{\text{thr}}$, min.\ iterations $L$, max.\ iterations $U$, increment $I$}
\KwOut{All evaluated morphologies $\mathcal{M}$}
Initialize population $P$ randomly\;
$\mathcal{M} \leftarrow P$; $evals \leftarrow 0$\;
\While{$evals < M$}{
  Train controller of each robot in $P$ for $L$ iterations\;
  $iters \leftarrow L$\;
  \Repeat{$r \leq r_{\text{thr}}$ \textbf{or} $iters \geq U$}{
    Select top $s \times N$ by fitness as survivors $\mathcal{S}$ \tcp*{pseudo selection}
    Estimate $\alpha$ for all robots in $P$ via Eq.\ (\ref{eq:core})\;
    $r \leftarrow \overline{\alpha}_{\mathcal{S}} \,/\, \overline{\alpha}_{P}$\;
    \lIf{$r > r_{\text{thr}}$ \textbf{and} $iters < U$}{extend all controllers by $I$ iterations; $iters \leftarrow iters + I$}
  }
  Mutate $\mathcal{S}$ to produce offspring $\mathcal{S}'$\;
  $P \leftarrow \mathcal{S} \cup \mathcal{S}'$; $\mathcal{M} \leftarrow \mathcal{M} \cup \mathcal{S}'$; $evals \leftarrow evals + |\mathcal{S}'|$\;
}
\Return{$\mathcal{M}$}\;
\end{algorithm}

\section{Experimental Study}
\label{sec:exp}
\subsection{Experimental Setup}
\label{sec:setup}
Our experiments are conducted on simulated VSRs in EvoGym \citep{bhatia2021evolution}. Following prior work \citep{bhatia2021evolution,liu2023rapidly,song2024morphvae,songlaser}, the VSR grid size is set to $5\times 5$, which already yields over $10^{17}$ possible morphologies, producing an expressive and challenging design space while remaining tractable for standard evolutionary algorithms \citep{mertan2025evolutionary}. We evaluate on three tasks: Carrier-v0, Pusher-v0, and BridgeWalker-v0 (Fig. \ref{Fig.1}(a)), spanning object manipulation and locomotion. The outer loop uses GA \citep{michalewicz2013genetic} for morphological evolution and the inner loop uses PPO \citep{schulman2017proximal} for control optimization, both canonical and widely adopted algorithms in the co-design literature \citep{bhatia2021evolution,song2024morphvae,zhao2025cross}, ensuring the generality of our findings.

To examine the impact of control complexity on morphological evolution, we vary the number of PPO iterations across five levels: 200, 500, 1000, and 2000 as \emph{weak} complexities, and 3000 as the \emph{strong} complexity, empirically found sufficient to reveal the true potential of most morphologies (we accordingly set $C^*=3000$). For brevity, we refer to the corresponding controllers as \emph{weak} and \emph{strong} controllers. For AdaControl, control complexity is dynamically adjusted per Section \ref{sec:method}. The minimal iterations $L$ is set to 500, half the commonly adopted setting \citep{bhatia2021evolution,song2024morphvae,songlaser}, allowing AdaControl to incrementally approach the appropriate complexity without excessive initial cost. The upper limit $U$ is set to 2000, which we find already largely eliminates selection bias while keeping computation manageable. The increment $I$ is set to 100. The MI ratio threshold $r_{\text{thr}}$ is set to 1.1, selected through the procedure described in Section \ref{sec:threshold}.

All experiments are allocated 1000 robot evaluations for fair comparison. Population size is 25, and the survival rate linearly decreases from 60\% to 8\% over the course of evolution, following \citet{bhatia2021evolution}. Results are averaged over five independent trials. All experiments are conducted on a server with Intel Xeon processors at 2.20 GHz without GPU acceleration.

In experiments with weak control and AdaControl, all evolved morphologies are re-evaluated using strong controllers to ensure fair performance comparison. This re-evaluation is performed solely for rigorous experimental analysis and is unnecessary in practical deployment.

Our experiments address the following questions:
\begin{itemize}
    \item \textbf{Q1:} To what extent does weak control complexity underestimate true potential?
    \item \textbf{Q2:} Does this underestimation bias evolutionary processes as conjectured in Section \ref{sec:theory}?
    \item \textbf{Q3:} How do such biases affect optimization efficiency and morphological diversity?
    \item \textbf{Q4:} Does AdaControl effectively resolve these issues?
\end{itemize}

\subsection{Evaluation Metrics}
We assess evolutionary processes with three metrics:
\begin{itemize}
    \item \textbf{Morphological intelligence (MI)}: the convergence speed of control learning, quantified as in Section \ref{sec:theory}. This metric is central to both our analysis of evolutionary bias and the dynamic scheduling in AdaControl. For comparative studies across control configurations, MI is computed from the complete learning curves of strong controllers, analogous to the re-evaluation of true potential described below. Within AdaControl, where complete curves are unavailable, MI is instead estimated from the learning curve segments observed up to the current iteration.
    \item \textbf{Maximal true potential}: the true potential of the best morphology found, evaluated with strong controllers (3000 PPO iterations), plotted against computational cost to yield performance curves reflecting optimization efficiency. We avoid the term \emph{fitness} to prevent ambiguity regarding which control complexity is used for evaluation.
    \item \textbf{Morphological diversity}: following \citet{saito2024effective}, we identify \emph{high-performing morphologies} as those exceeding the top $k\%$ quantile of true potential across all experiments ($k=5$, following \citet{song2024morphvae}), and report the average pairwise edit distance among them. Diversity reflects a co-design system's capacity to discover varied capable designs \citep{medvet2021biodiversity,pigozzi2023factors}.
\end{itemize}

\subsection{Baselines}
Beyond GA with various fixed control complexities, we compare against three additional baselines. \textbf{MorphVAE} \citep{song2024morphvae} and \textbf{LASeR} \citep{songlaser} are state-of-the-art VSR co-design algorithms based on deep generative models. MorphVAE fits a variational autoencoder (VAE) to the distribution of high-performing morphologies and samples new candidates from the learned latent space, with the VAE iteratively updated after each round of natural selection. LASeR replaces the VAE with a pre-trained large language model, leveraging its in-context learning and generation capabilities to fit and sample morphologies. \textbf{FitControl} adapts the self-adaptive learning cycle of \citet{le2024improving}, which schedules per-morphology training duration based on a target fitness improvement $\delta$. We adopt this scheduling mechanism within our synchronous population-based framework for controlled comparison.

\subsection{Impact of Control Complexity on Evolutionary Bias}
\label{sec:bias}
\subsubsection{Underestimation of True Potential}
We first validate that 3000 PPO iterations suffices to accurately estimate true potential. In the strong-control experiments, learning curves converge well before this budget, with average convergence iterations of 2200.78, 2216.28, and 1731.16 for Carrier, Pusher, and BridgeWalker, respectively (Table \ref{tab:convergence_stats}, last column). Here, a learning curve is deemed converged when performance first reaches within 5\% of its peak value $f_{\text{max}}$. We therefore adopt 3000 iterations as the strong control complexity, serving as the ground-truth reference for evaluating true potential and benchmarking all other control configurations.

\begin{table}[htbp]
\centering
\caption{Mean and standard deviation of PPO convergence iterations under different control complexities.}
\label{tab:convergence_stats}
\resizebox{\textwidth}{!}{%
\renewcommand{\arraystretch}{1.3}
\begin{tabular}{lccccc}
\toprule
\textbf{Task} & \multicolumn{5}{c}{\textbf{Control Complexity}} \\
\cline{2-6}
 & \textbf{200} & \textbf{500} & \textbf{1000} & \textbf{2000} & \textbf{3000} \\
\midrule
\textbf{Carrier} & 2086.40 $\pm$ 743.67 & 1995.46 $\pm$ 802.65 & 2064.89 $\pm$ 768.64 & 2116.82 $\pm$ 735.35 & 2200.78 $\pm$ 674.62 \\
\textbf{Pusher} & 2094.57 $\pm$ 685.21 & 2080.44 $\pm$ 708.53 & 2141.01 $\pm$ 679.03 & 2145.09 $\pm$ 646.19 & 2216.28 $\pm$ 634.45 \\
\textbf{BridgeWalker} & 1633.38 $\pm$ 855.06 & 1481.07 $\pm$ 852.35 & 1589.40 $\pm$ 843.95 & 1691.15 $\pm$ 820.79 & 1731.16 $\pm$ 819.92 \\
\bottomrule
\end{tabular}%
}
\end{table}

\begin{figure}[h]
    \centering
    \begin{subfigure}{0.5\textwidth}
        \includegraphics[width=\textwidth]{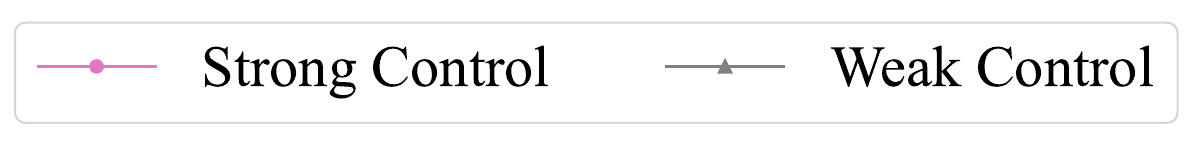}
    \end{subfigure}

    \centering
    \begin{subfigure}{0.42\textwidth}
        \includegraphics[width=\textwidth]{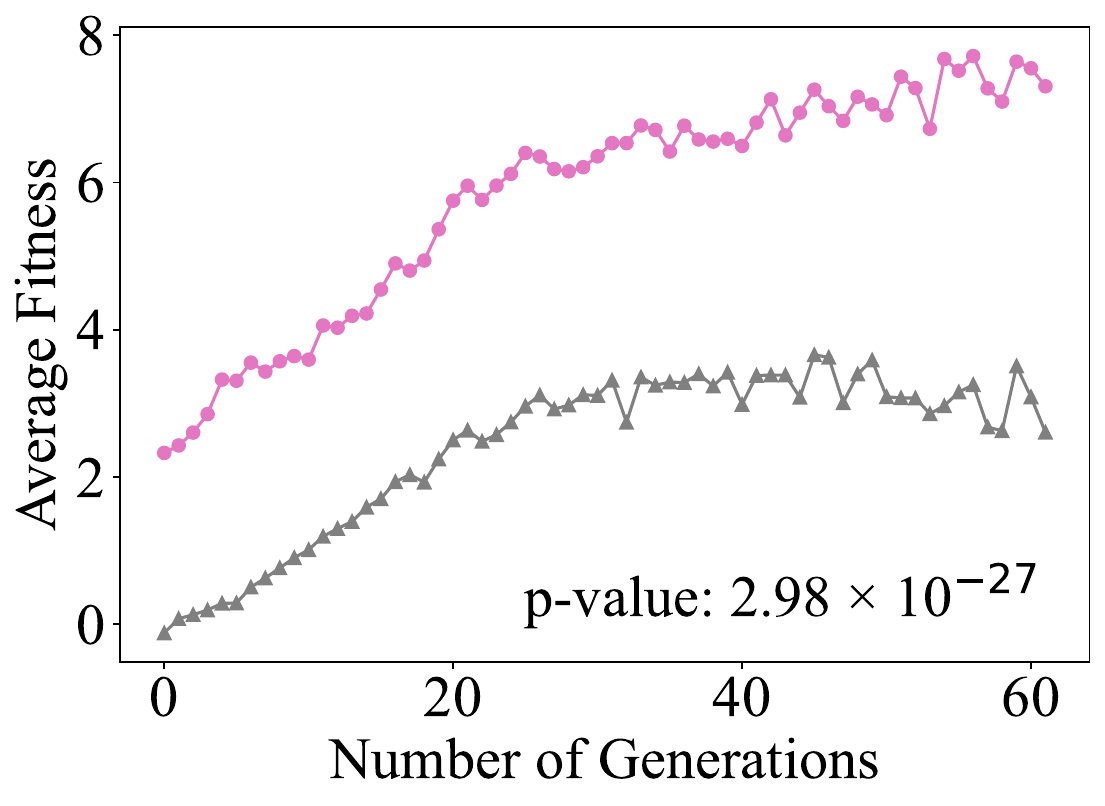}
        \caption{Control complexity = 200}
        \label{fig:sub1}
    \end{subfigure}
    \hfill
    \begin{subfigure}{0.42\textwidth}
        \includegraphics[width=\textwidth]{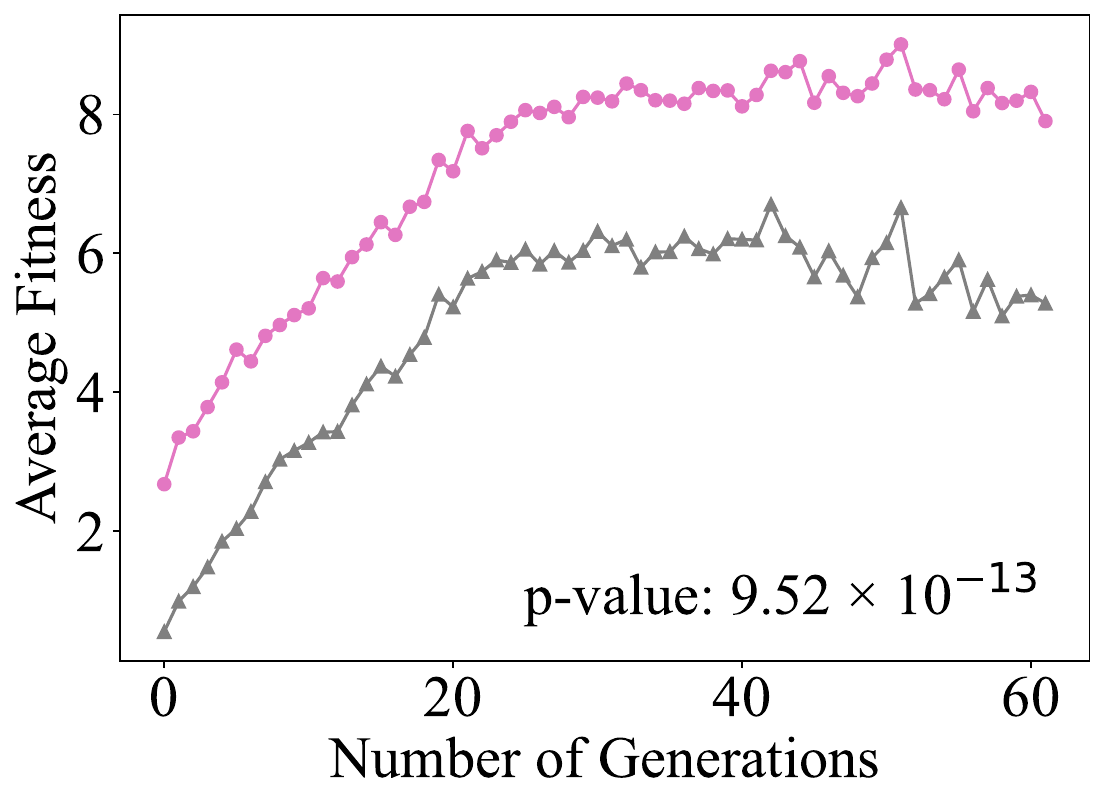}
        \caption{Control complexity = 500}
        \label{fig:sub2}
    \end{subfigure}

    \vspace{0.3cm}

    \begin{subfigure}{0.42\textwidth}
        \includegraphics[width=\textwidth]{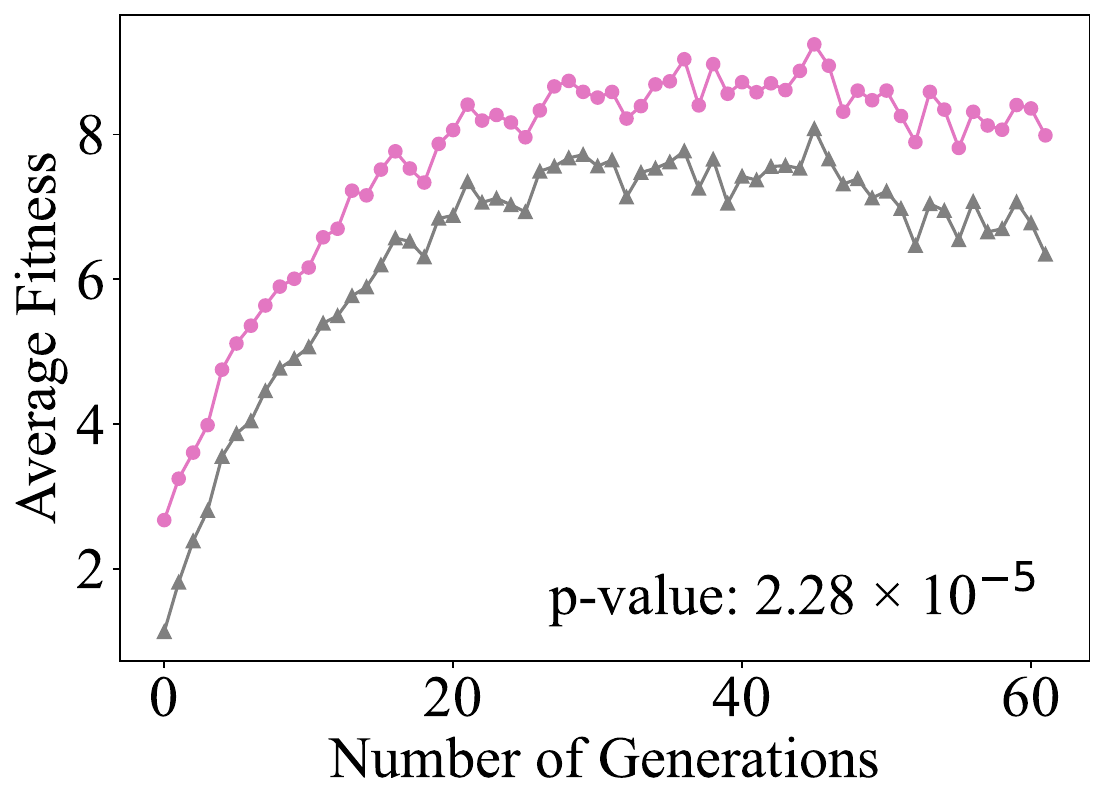}
        \caption{Control complexity = 1000}
        \label{fig:sub3}
    \end{subfigure}
    \hfill
    \begin{subfigure}{0.42\textwidth}
        \includegraphics[width=\textwidth]{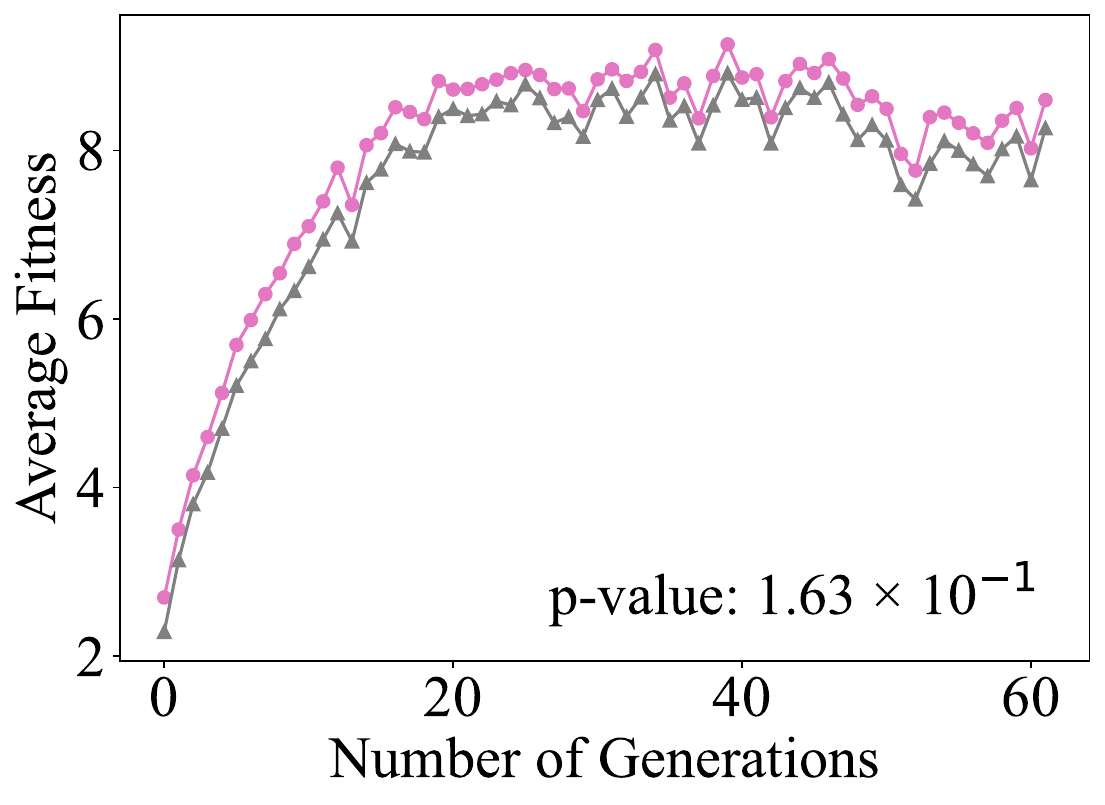}
        \caption{Control complexity = 2000}
        \label{fig:sub4}
    \end{subfigure}

    \caption{True potential versus weak-control-evaluated fitness under varying control complexities in Carrier. Annotated $p$-values are from two-tailed $t$-tests.}
    \label{fig:underestimation}
\end{figure}

We now investigate whether and to what extent weak control underestimates true potential (\textbf{Q1}). For each weak complexity, we run co-design experiments and compare the population averages of weak-control-evaluated fitness against true potential. Fig. \ref{fig:underestimation} reports Carrier results with two-tailed $t$-test $p$-values annotated. Pusher and BridgeWalker yield qualitatively consistent results and are omitted for brevity. Two findings emerge: (a) weak controllers consistently underestimate true potential, with statistically significant discrepancies at 200, 500, and 1000 iterations; (b) the gap narrows with increasing control complexity and becomes insignificant at 2000 iterations. These results confirm that the underestimation, while pronounced under weak control, can be effectively alleviated by extending control learning.

\subsubsection{Evolutionary Biases of Weak Controllers}
\begin{figure*}[h]
    \centering
    \begin{subfigure}{0.3\textwidth}
        \includegraphics[width=\textwidth]{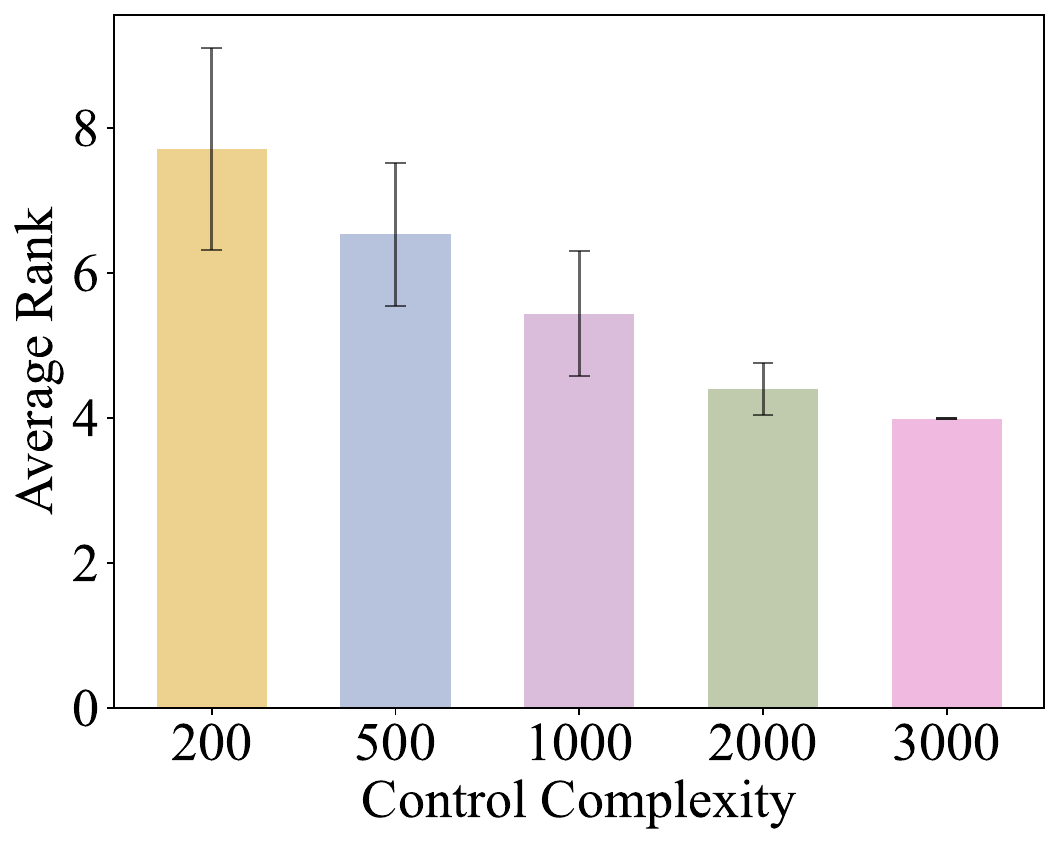}
        \caption{Carrier}
    \end{subfigure}
    \hfill
    \begin{subfigure}{0.3\textwidth}
        \includegraphics[width=\textwidth]{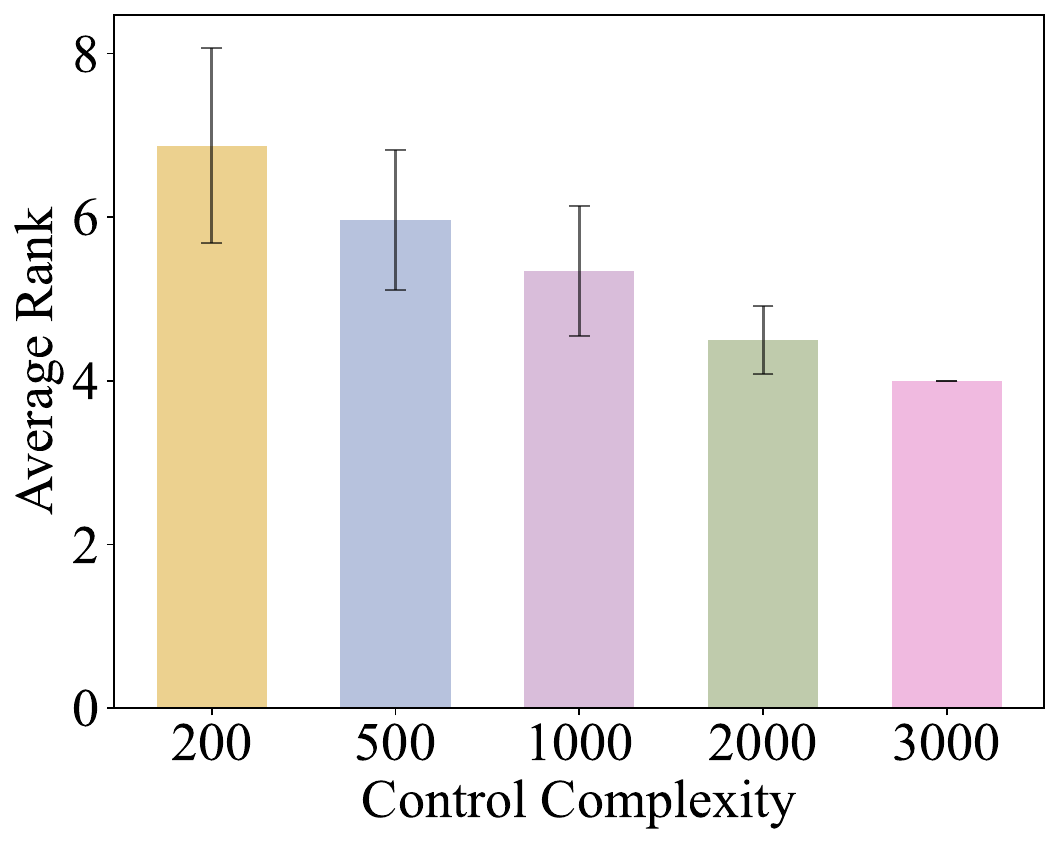}
        \caption{Pusher}
    \end{subfigure}
    \hfill
    \begin{subfigure}{0.3\textwidth}
        \includegraphics[width=\textwidth]{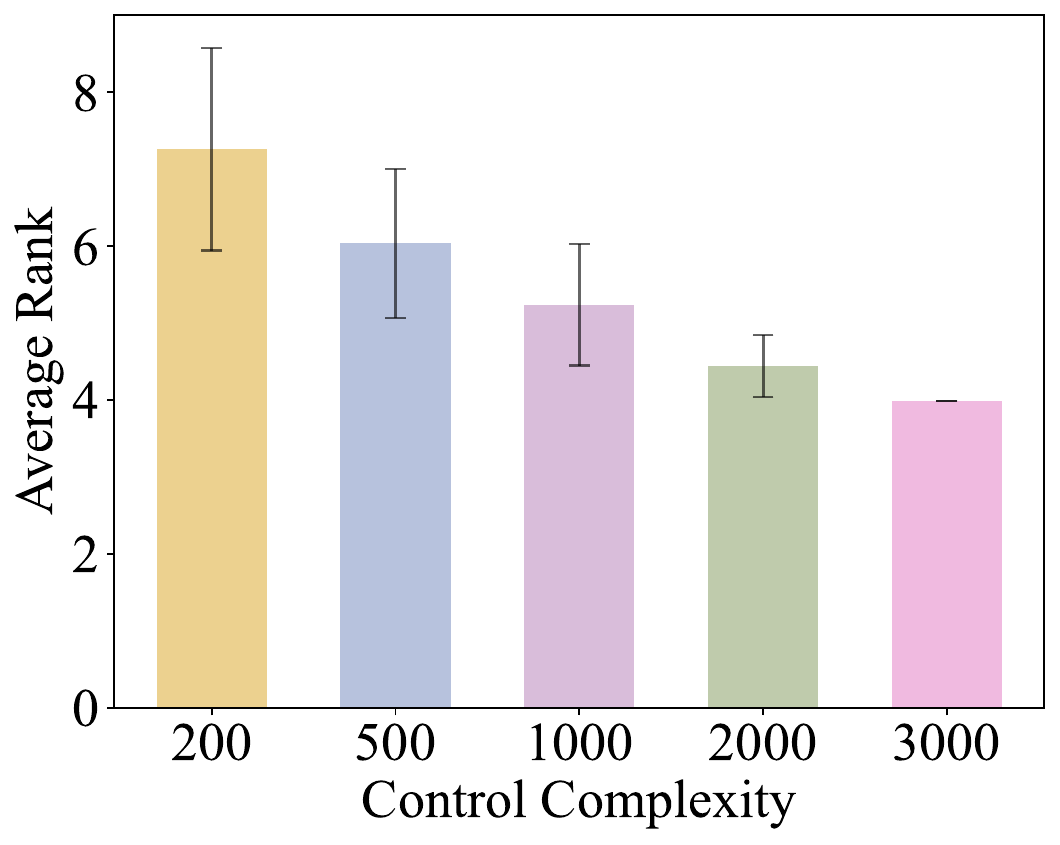}
        \caption{BridgeWalker}
    \end{subfigure}

    \caption{Average ranking of survivors when re-sorted by true potential rather than weak-control fitness. Lower rankings indicate greater deviation from true-potential-based selection.}
    \label{fig:ranking}
\end{figure*}

\begin{figure*}[h]
    \centering
    \begin{subfigure}{0.5\textwidth}
        \includegraphics[width=\textwidth]{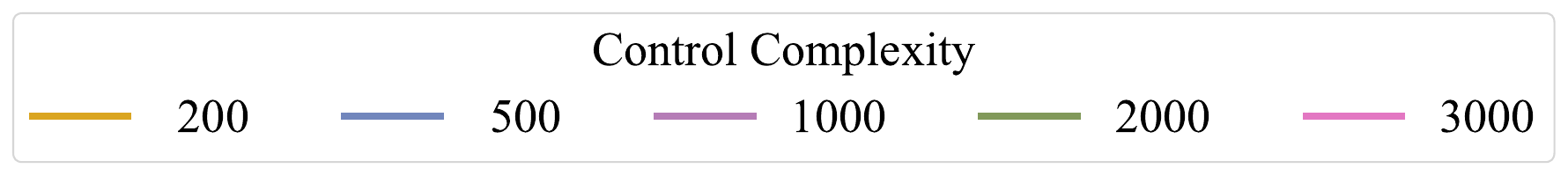}
    \end{subfigure}

    \centering
    \begin{subfigure}{0.3\textwidth}
        \includegraphics[width=\textwidth]{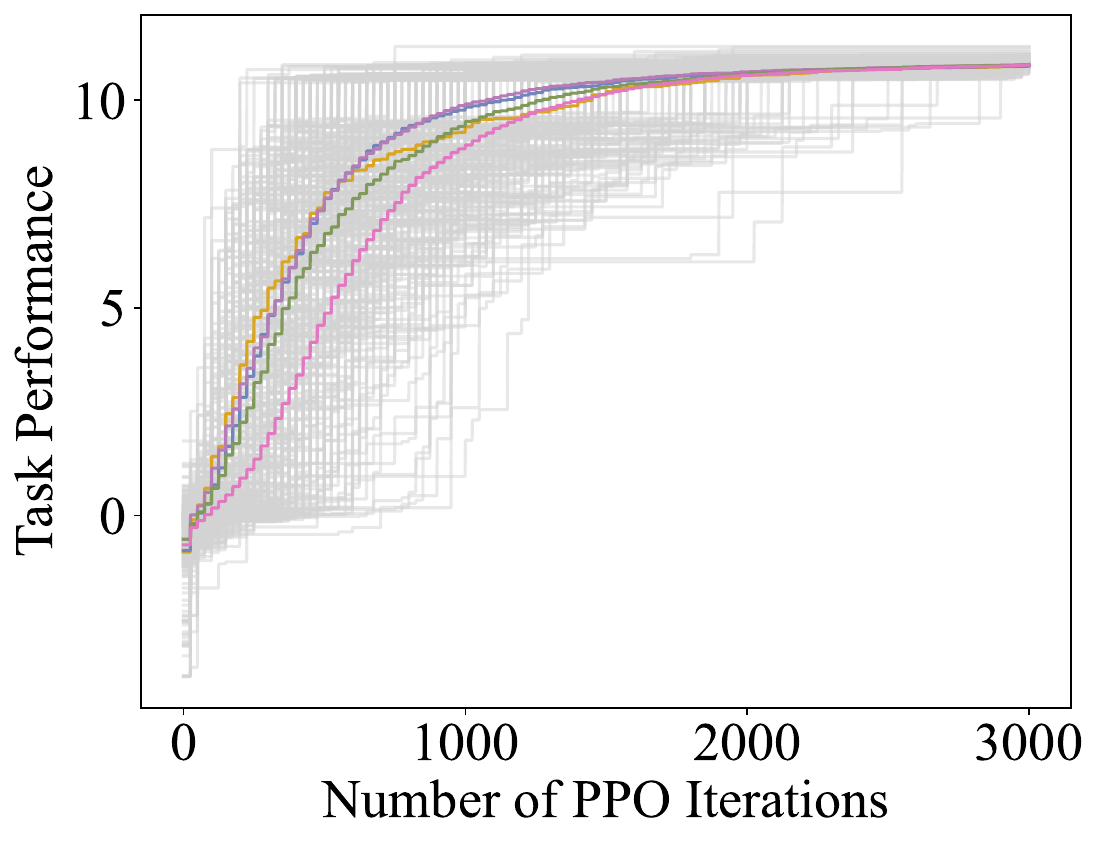}
        \caption{Carrier}
    \end{subfigure}
    \hfill
    \begin{subfigure}{0.3\textwidth}
        \includegraphics[width=\textwidth]{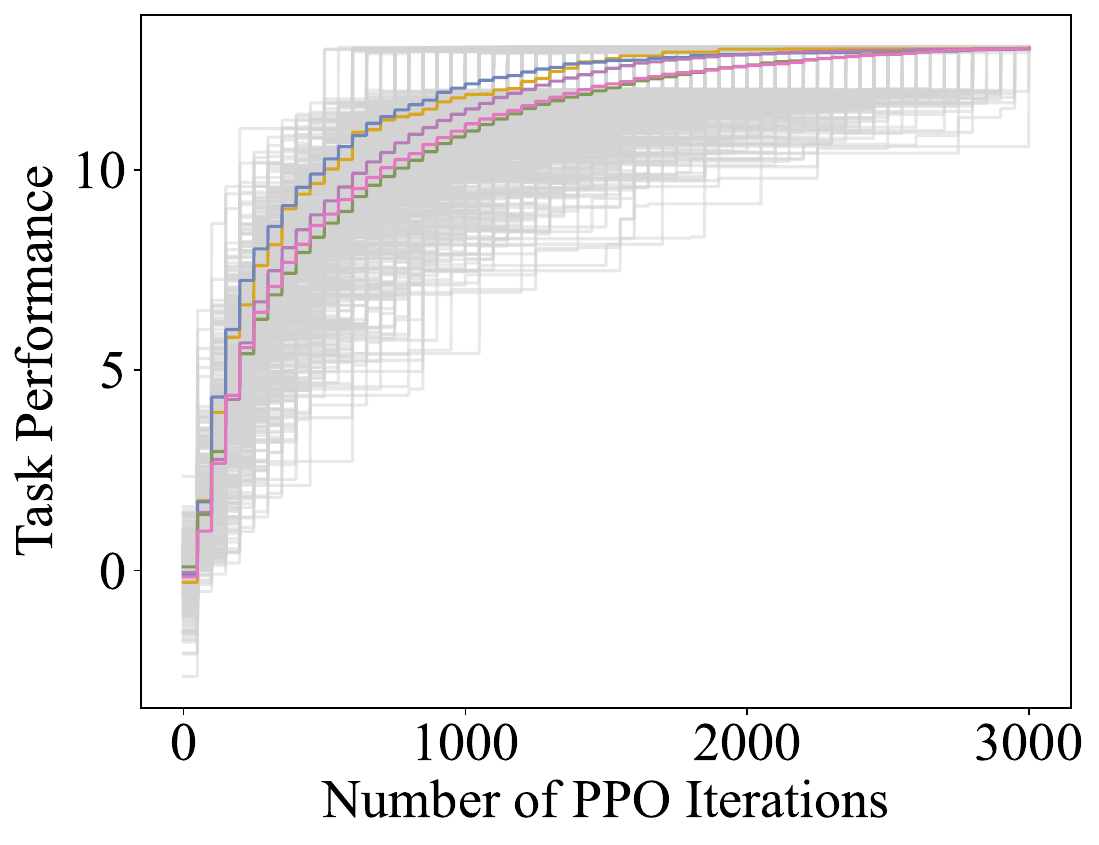}
        \caption{Pusher}
    \end{subfigure}
    \hfill
    \begin{subfigure}{0.3\textwidth}
        \includegraphics[width=\textwidth]{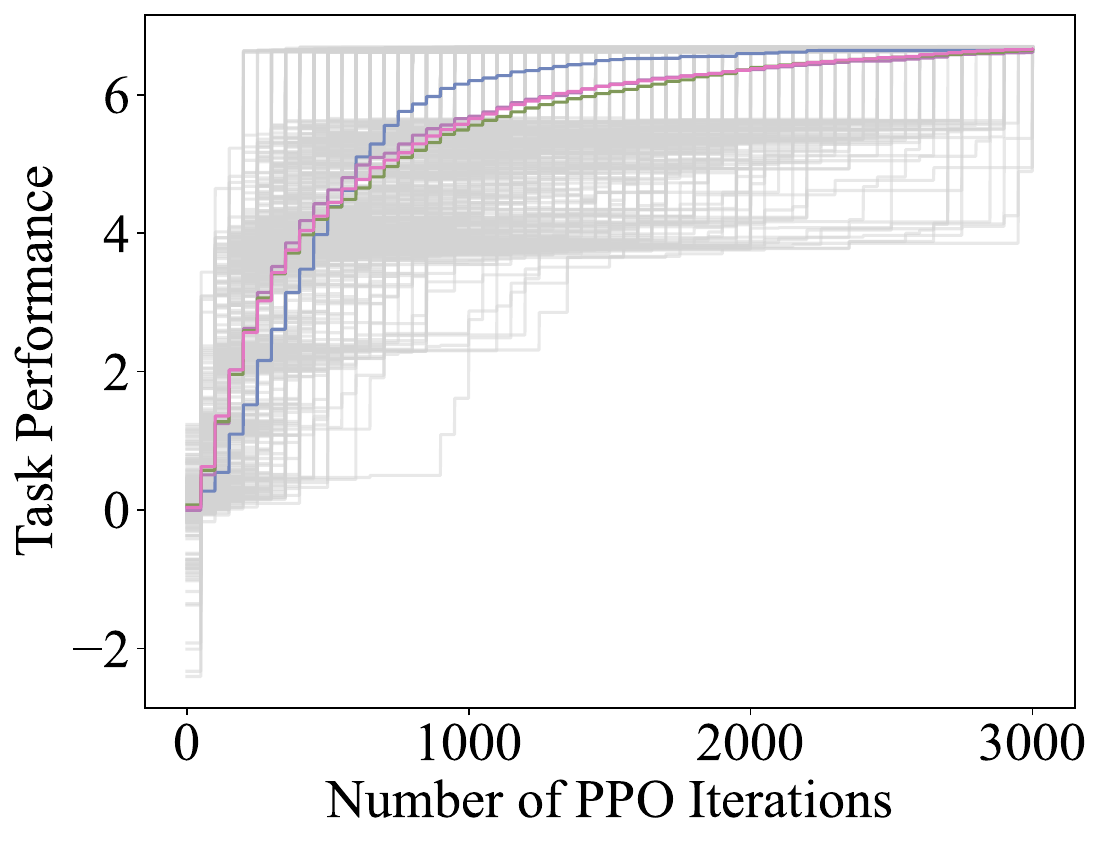}
        \caption{BridgeWalker}
    \end{subfigure}

    \caption{Learning curves of high-performing morphologies evolved under different control complexities. Individual curves in gray; colored curves show the average per complexity. }
    \label{fig:curves}
\end{figure*}

\begin{figure*}[h]\centering
    \begin{subfigure}{0.3\textwidth}
        \includegraphics[width=\textwidth]{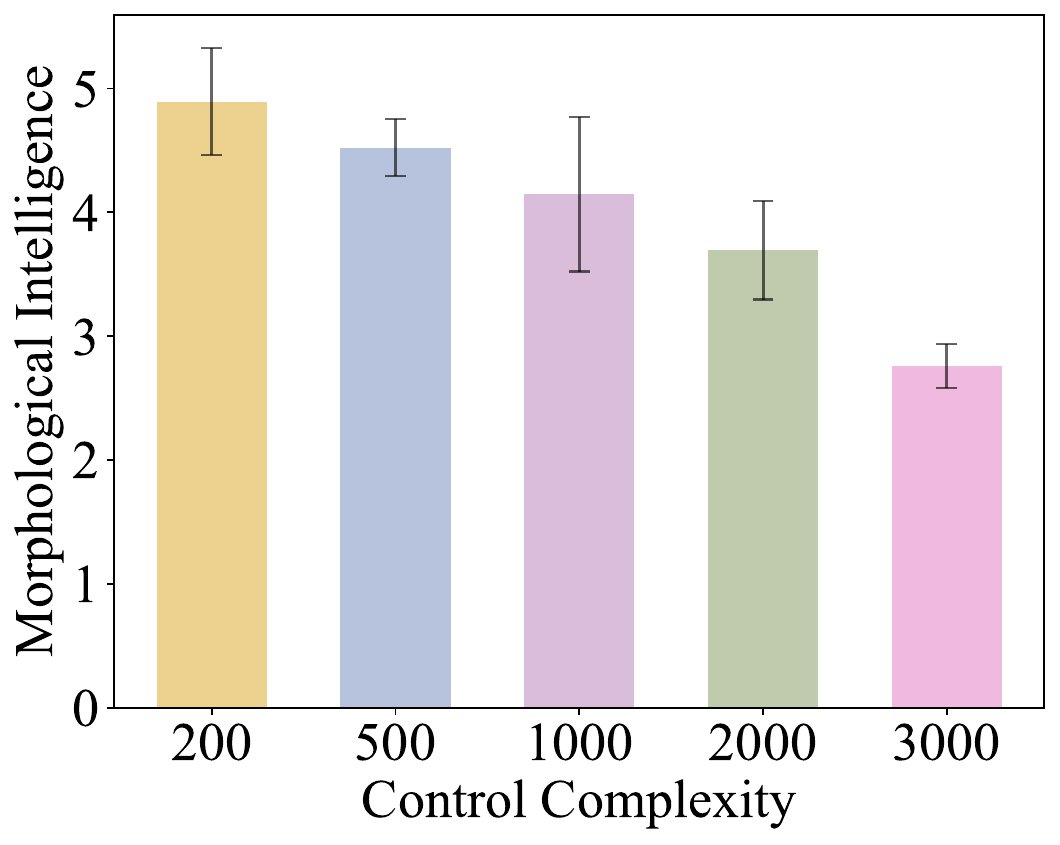}
        \caption{Carrier}
    \end{subfigure}
    \hfill
    \begin{subfigure}{0.3\textwidth}
        \includegraphics[width=\textwidth]{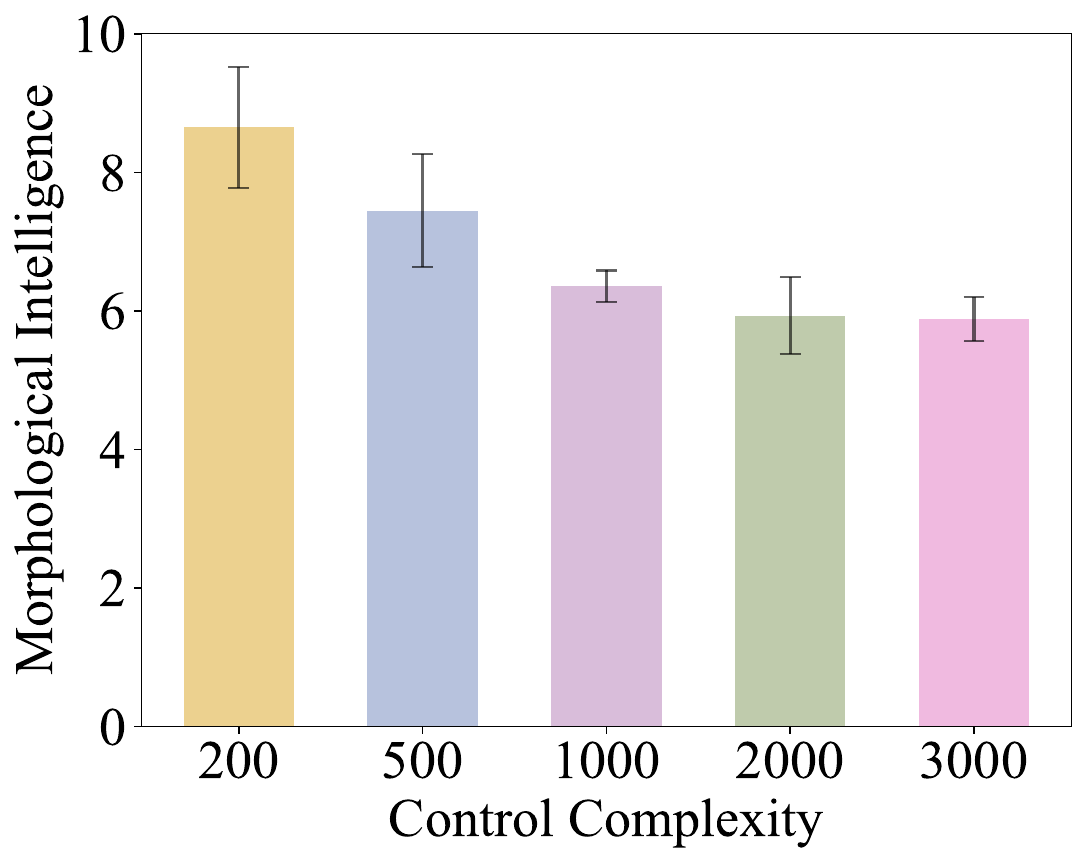}
        \caption{Pusher}
    \end{subfigure}
    \hfill
    \begin{subfigure}{0.3\textwidth}
        \includegraphics[width=\textwidth]{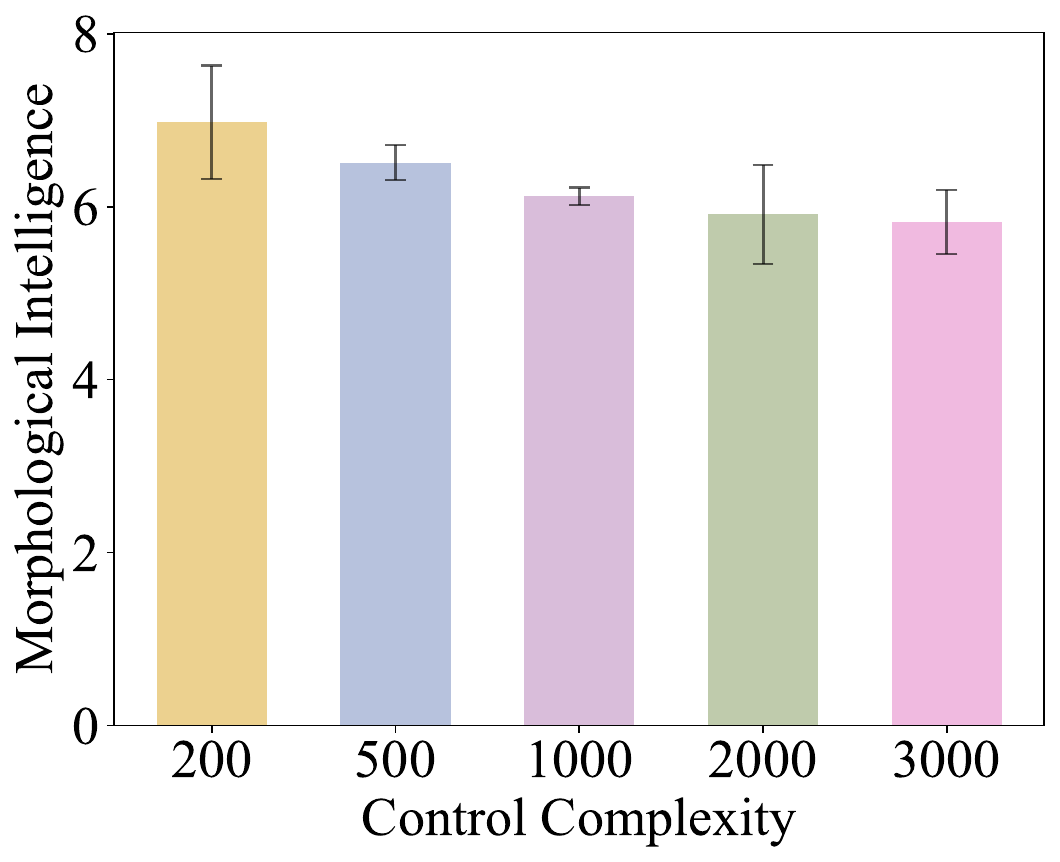}
        \caption{BridgeWalker}
    \end{subfigure}

    \caption{Average MI of all evaluated morphologies throughout evolution under different control complexities.}
    \label{fig:intelligence}
\end{figure*}

The preceding results show that weak control systematically underestimates true potential. A natural follow-up is whether this underestimation translates into biased natural selection (\textbf{Q2}). To test this, we re-rank survivors by true potential rather than weak-control fitness. As shown in Fig. \ref{fig:ranking}, weaker controllers produce survivors that rank progressively lower in true potential, confirming that selection increasingly deviates from true morphological quality. Fig. \ref{fig:curves} further compares the learning curves of high-performing morphologies evolved under different control complexities. Individual curves are plotted in gray, with colored curves showing the average for each complexity. Weaker controllers consistently select morphologies with steeper initial learning curves, while stronger controllers (2000 and 3000 iterations) yield gentler average curves, indicating greater tolerance for slow but ultimately superior learners. This pattern is quantitatively confirmed in Table \ref{tab:convergence_stats} and Fig. \ref{fig:intelligence}: morphologies evolved under weaker control converge faster and exhibit systematically higher MI.

These results answer \textbf{Q2} affirmatively: prematurely terminated control learning biases natural selection towards fast learners and substantially shapes evolutionary outcomes. Strong control, by contrast, more faithfully evaluates true potential and thereby preserves slow but capable morphologies. Notably, the \emph{morphological Baldwin effect} reported in \citet{gupta2021embodied} emerges as a special case of this bias rather than a general evolutionary phenomenon, arising specifically when control learning is insufficient to distinguish fast learners from good ones.

\subsection{Impact of Evolutionary Bias on Co-Design Performance}
In this section, we examine how the evolutionary biases identified above affect co-design performance (\textbf{Q3}) and evaluate AdaControl as a remedy (\textbf{Q4}). We analyze three dimensions: morphological intelligence dynamics, optimization efficiency, and morphological diversity.

\subsubsection{Morphological Intelligence}
\begin{figure*}[h]
    \centering
    \begin{subfigure}{0.65\textwidth}
        \includegraphics[width=\textwidth]{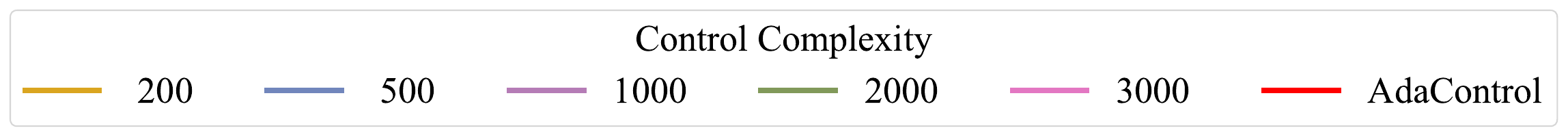}
    \end{subfigure}

    \centering
    \begin{subfigure}{0.3\textwidth}
        \includegraphics[width=\textwidth]{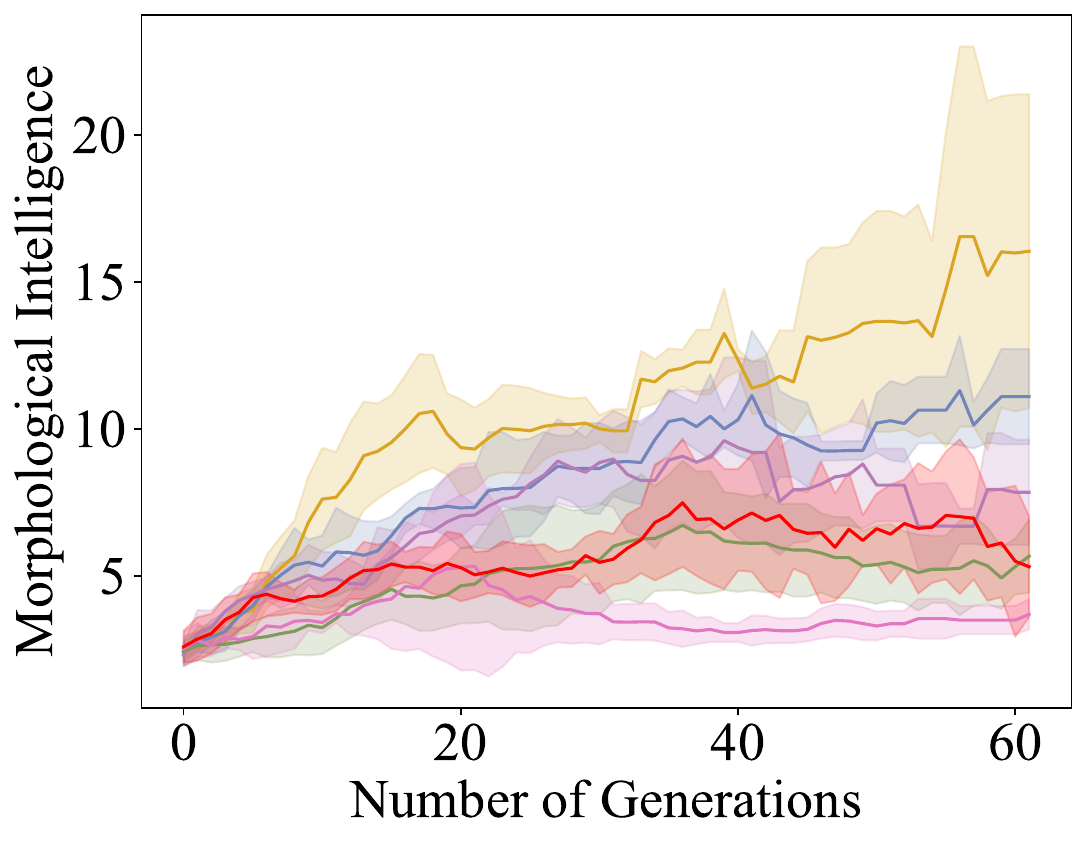}
        \caption{Carrier}
    \end{subfigure}
    \hfill
    \begin{subfigure}{0.3\textwidth}
        \includegraphics[width=\textwidth]{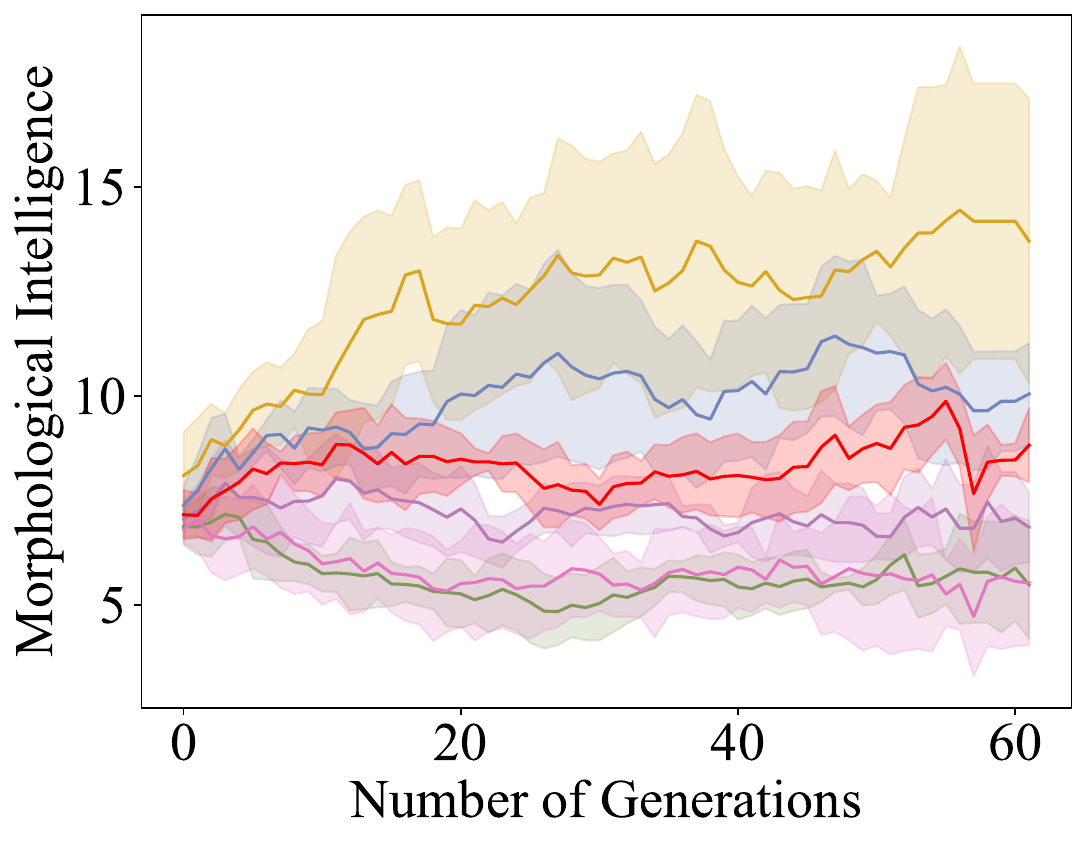}
        \caption{Pusher}
    \end{subfigure}
    \hfill
    \begin{subfigure}{0.3\textwidth}
        \includegraphics[width=\textwidth]{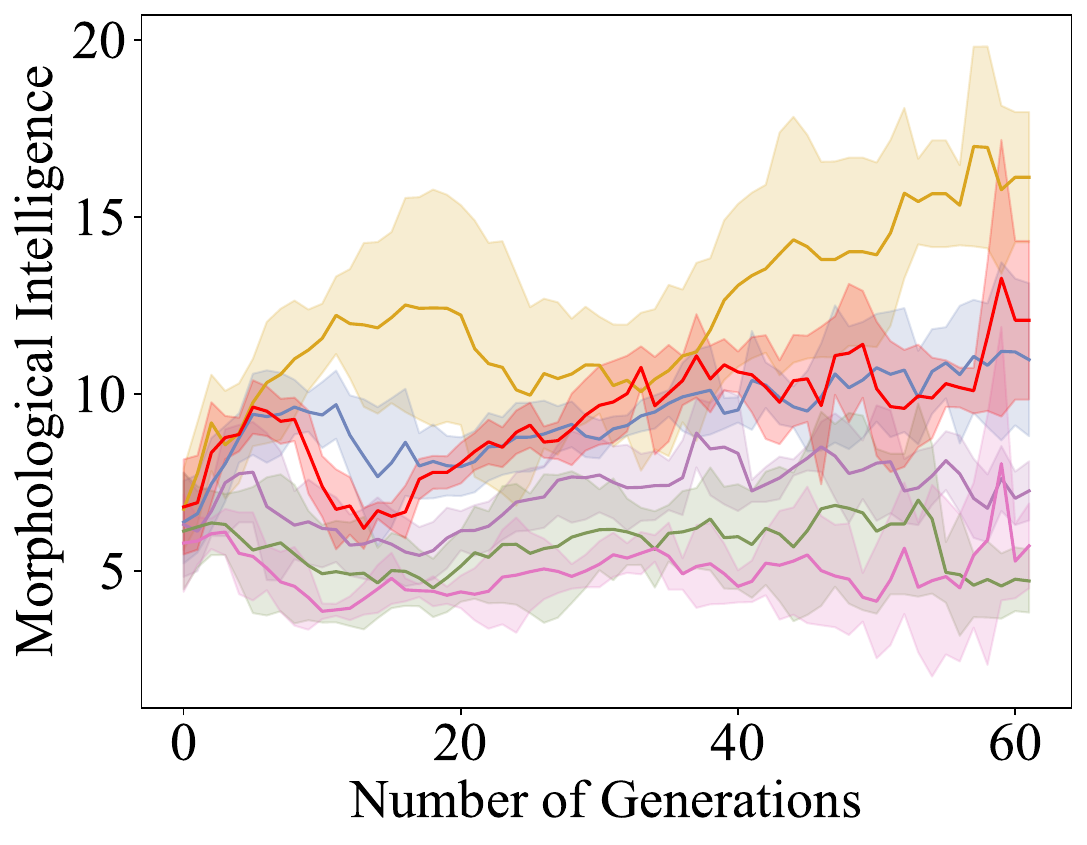}
        \caption{BridgeWalker}
    \end{subfigure}

    \caption{Population-average MI of survivors per generation. Shaded areas indicate standard deviation. Weak control produces a rising MI trend, signaling accumulating bias; strong control and AdaControl maintain stable MI.}
    \label{fig:trace}
\end{figure*}

Section \ref{sec:bias} established that weak control biases selection towards higher MI. We now examine how this bias evolves across generations. Fig. \ref{fig:trace} plots the population-average MI of survivors against generation number. Under weak control, MI exhibits a clear upward trend, most pronounced at 200 iterations, indicating that the preference for fast learners compounds over successive generations. Strong control maintains MI at a stable, moderate level. AdaControl substantially reduces MI growth relative to weak control across all three tasks, with near-complete stabilization in Carrier and Pusher and a more moderate effect in BridgeWalker, while using far fewer PPO iterations than strong control (Table \ref{tab:ppo_compare}).

The more pronounced MI growth in BridgeWalker raises the question of whether $r_{\text{thr}}=1.1$ is overly lenient for this task. However, as shown in Section \ref{sec:threshold}, tightening the threshold does not improve design space coverage, leading us to identify two competing factors simultaneously governed by $r_{\text{thr}}$.

\begin{table}[h]
\centering
\footnotesize
\renewcommand{\arraystretch}{1.3}
\caption{Average PPO iterations per evaluation for AdaControl and FitControl (mean $\pm$ std). AdaControl reduces control learning cost by 65--81\% relative to strong control (3000 iterations) while achieving comparable or superior optimization efficiency and morphological diversity.}
\label{tab:ppo_compare}
\begin{tabular}{lcc}
\toprule
\textbf{Task} & \textbf{AdaControl} & \textbf{FitControl} \\
\midrule
Carrier & 1063.62 $\pm$ 194.48 & 1631.64 $\pm$ 324.59 \\
Pusher & 609.82 $\pm$ 124.62 & 1959.40 $\pm$ 211.11 \\
BridgeWalker & 584.00 $\pm$ 62.29 & 1443.45 $\pm$ 81.24 \\
\bottomrule
\end{tabular}
\end{table}

\subsubsection{Optimization Efficiency}
\begin{figure*}[h]
    \centering
    \begin{subfigure}{0.32\textwidth}
        \includegraphics[width=\textwidth]{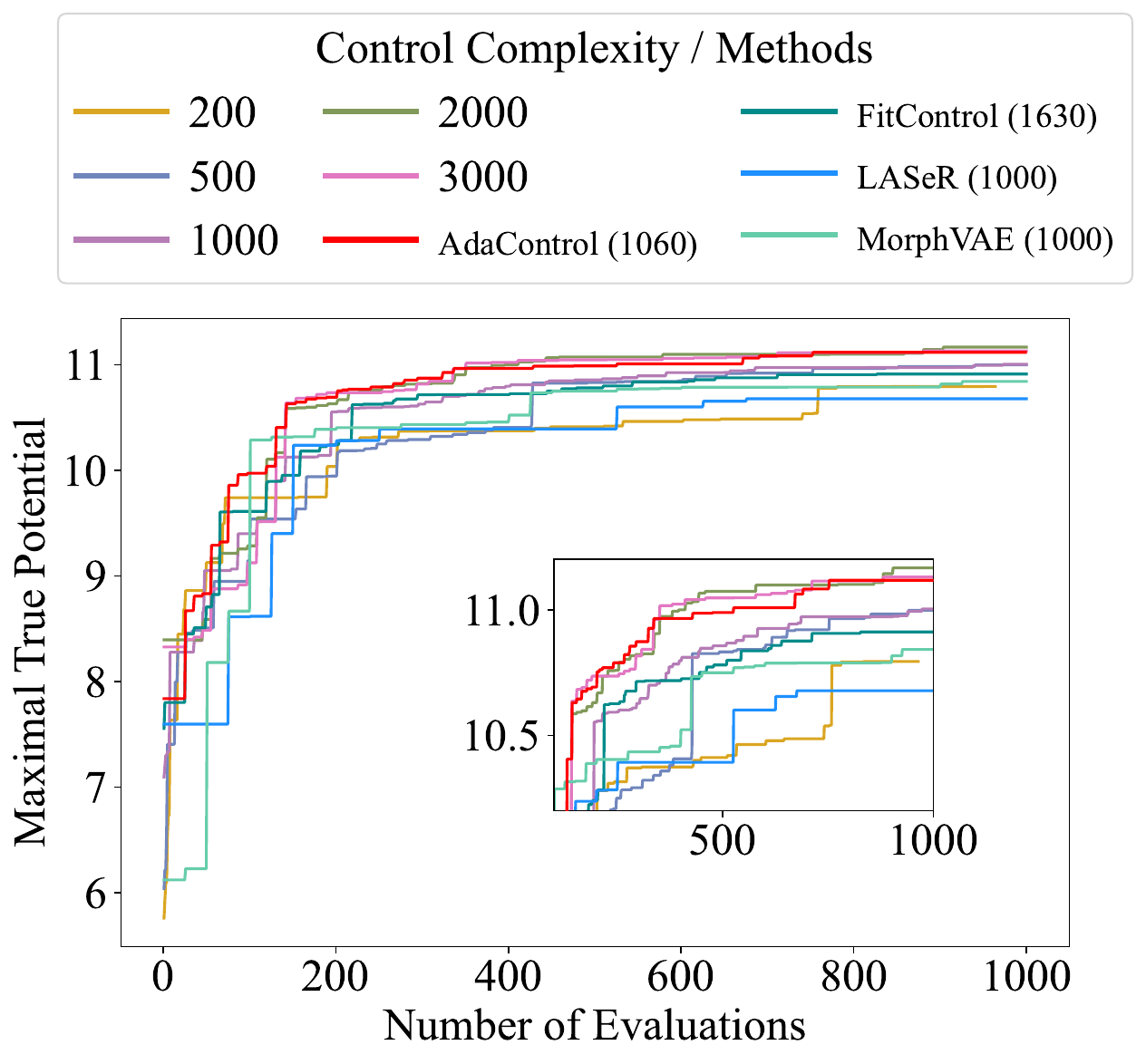}
        \caption{Carrier}
    \end{subfigure}
    \hfill
    \begin{subfigure}{0.32\textwidth}
        \includegraphics[width=\textwidth]{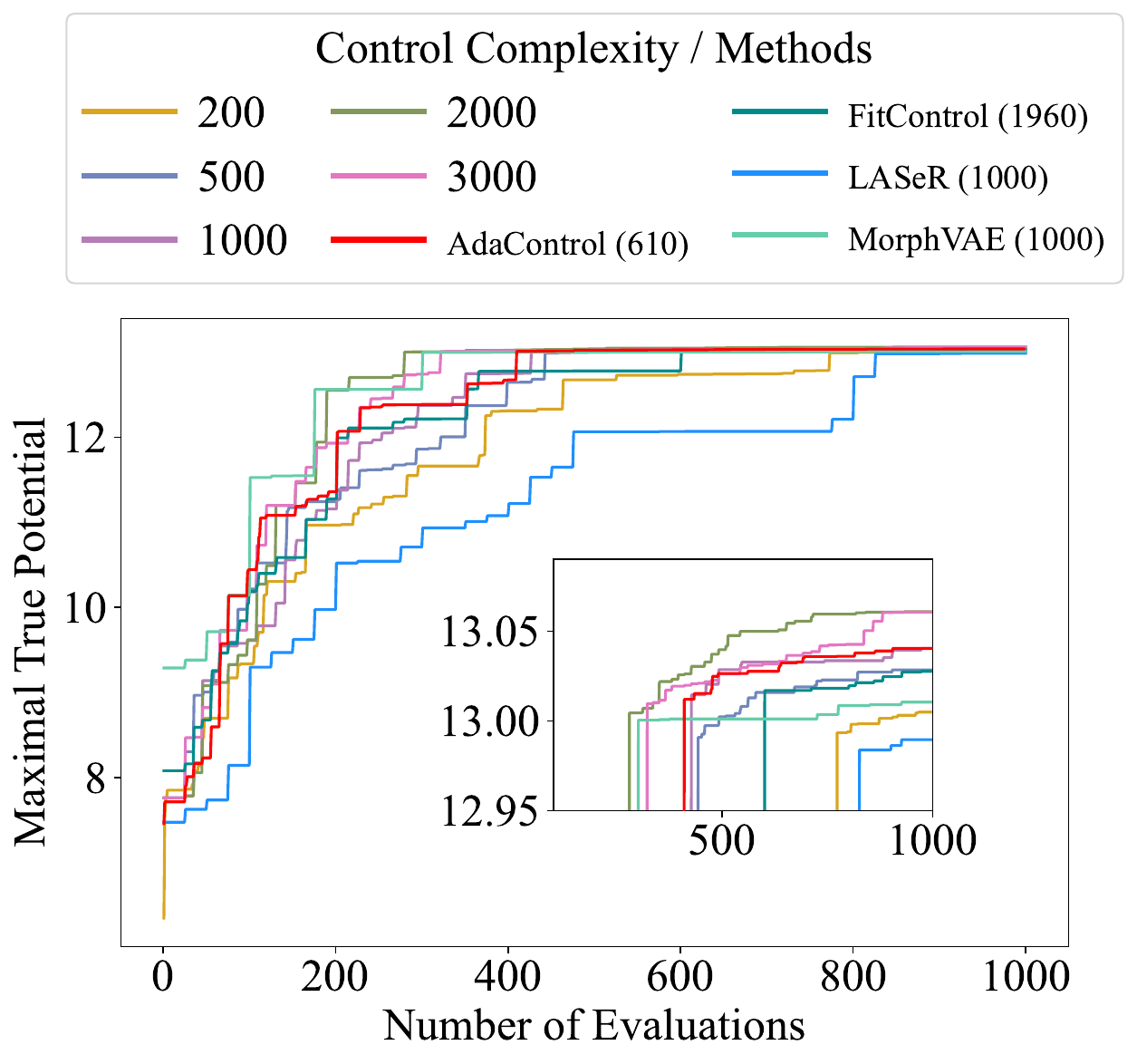}
        \caption{Pusher}
    \end{subfigure}
    \hfill
    \begin{subfigure}{0.32\textwidth}
        \includegraphics[width=\textwidth]{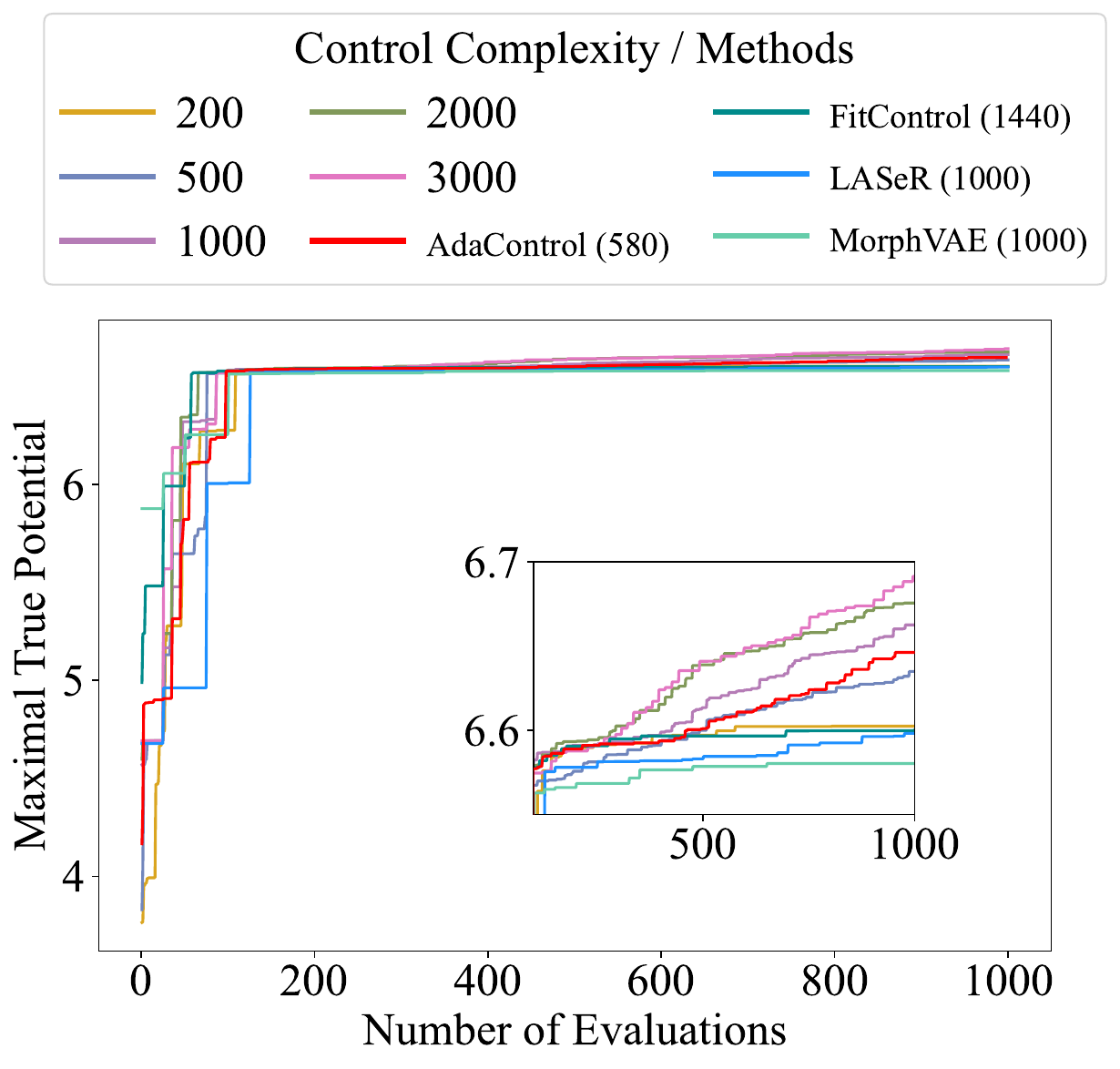}
        \caption{BridgeWalker}
    \end{subfigure}

    \caption{Optimization efficiency measured by number of robot evaluations. Numbers in parentheses denote average PPO iterations. Insets magnify the later stages of evolution.}
    \label{fig:efficiency}
\end{figure*}

\begin{figure*}[h]
    \centering

\begin{subfigure}{0.7\textwidth}
        \includegraphics[width=\textwidth]{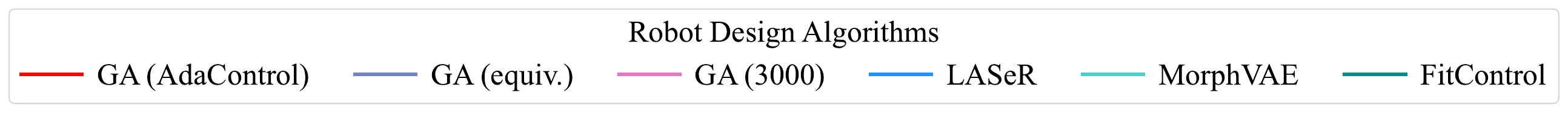}
    \end{subfigure}

    \begin{subfigure}{0.32\textwidth}
        \includegraphics[width=\textwidth]{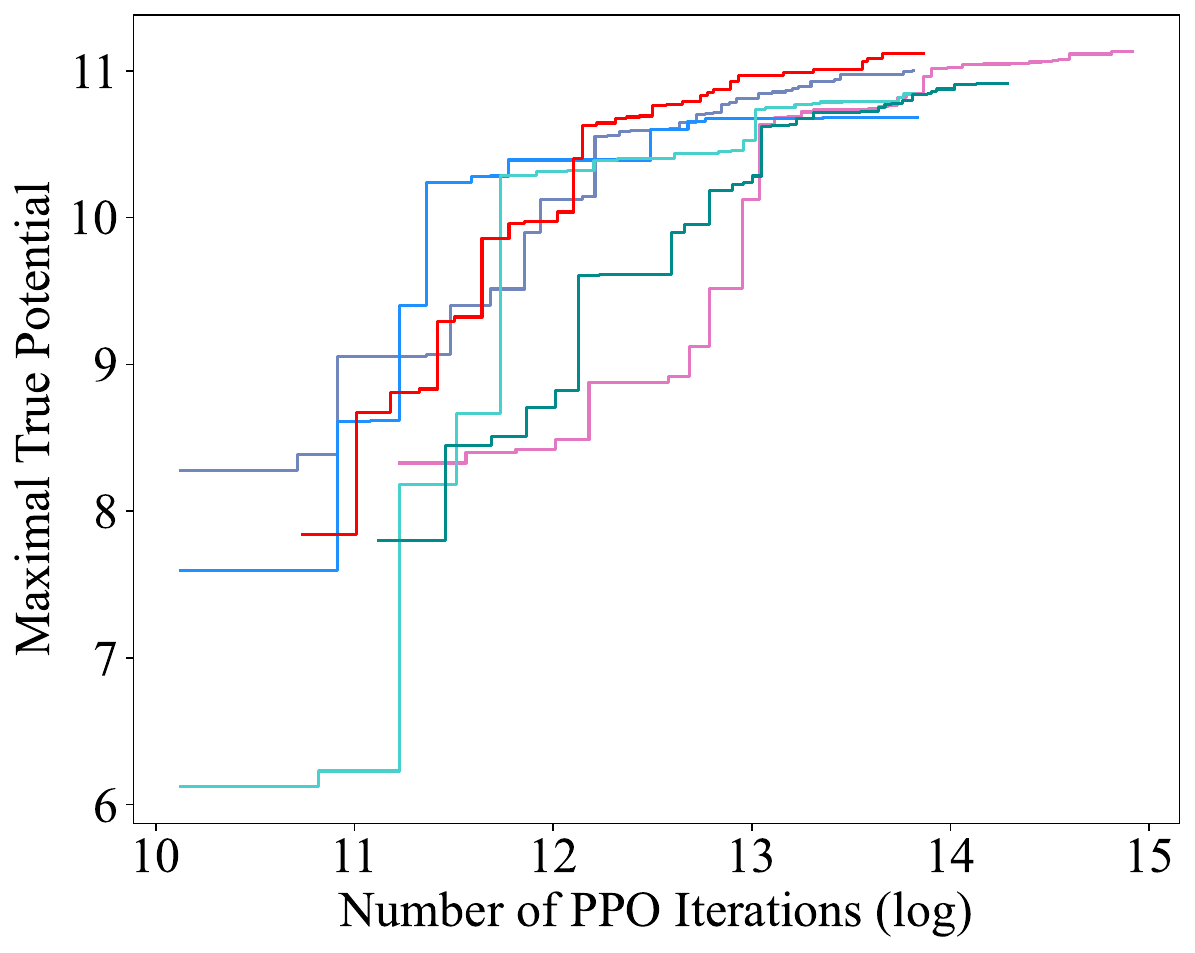}
        \caption{Carrier}
    \end{subfigure}
    \hfill
    \begin{subfigure}{0.32\textwidth}
        \includegraphics[width=\textwidth]{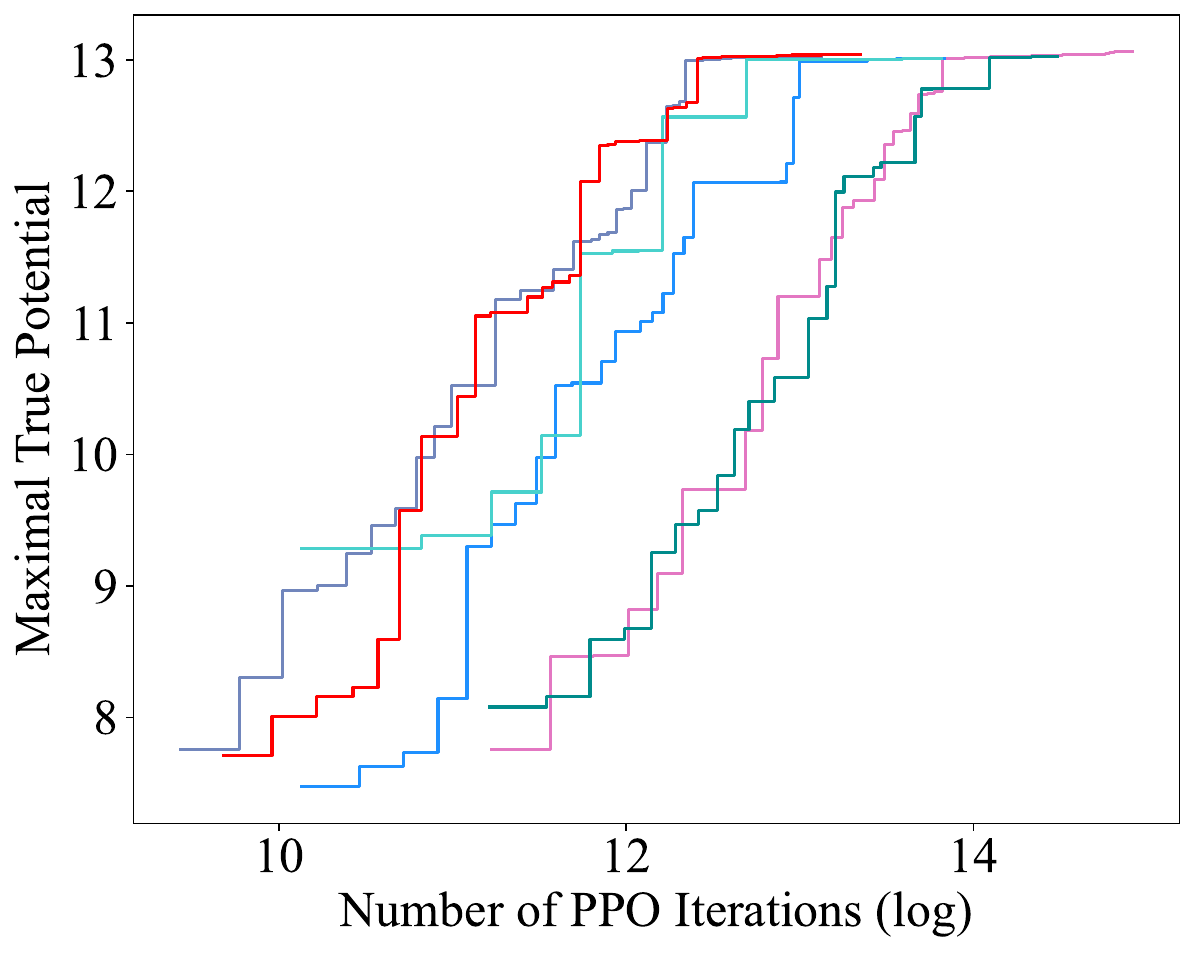}
        \caption{Pusher}
    \end{subfigure}
    \hfill
    \begin{subfigure}{0.32\textwidth}
        \includegraphics[width=\textwidth]{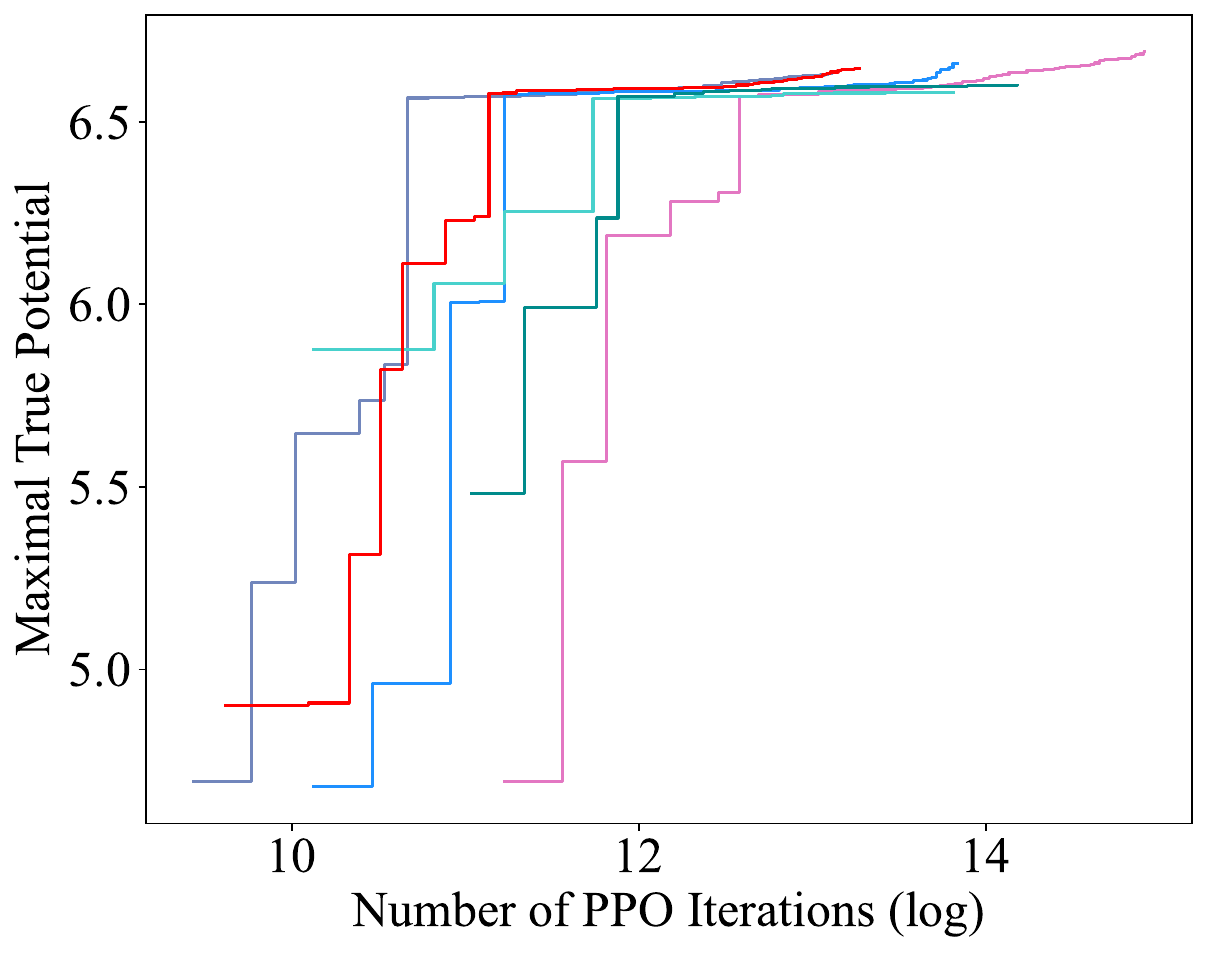}
        \caption{BridgeWalker}
    \end{subfigure}

    \caption{Optimization efficiency measured by cumulative PPO iterations (log scale). ``equiv.'' denotes the fixed complexity approximately matching AdaControl's average iterations (1000 for Carrier, 500 for Pusher and BridgeWalker).}
    \label{fig:iters}
\end{figure*}

We assess optimization efficiency by plotting the best true potential found against the number of robot evaluations (Fig. \ref{fig:efficiency}) and against cumulative PPO iterations on a logarithmic scale (Fig. \ref{fig:iters}). Note that MorphVAE and LASeR follow their original experimental settings with 1000 PPO iterations for control learning.

In terms of evaluation count (Fig. \ref{fig:efficiency}), higher control complexity generally yields better efficiency, as unbiased fitness evaluation enables more thorough exploration and avoidance of local optima. AdaControl achieves performance comparable to strong control across all three tasks while using far fewer PPO iterations per evaluation. FitControl is competitive in Carrier and Pusher but offers less consistent gains in BridgeWalker. The diminishing gap between 2000 and 3000 iterations suggests that simply prolonging control learning faces diminishing returns.

The advantage of adaptive scheduling is more evident when efficiency is measured by cumulative control learning cost (Fig. \ref{fig:iters}). For the same iteration budget, AdaControl discovers substantially higher-performing morphologies than strong control, or equivalently, reaches comparable performance at a fraction of the cost. This advantage arises from AdaControl's flexible allocation of control resources, which concentrates additional learning on generations where selection bias is detected rather than distributing iterations uniformly across all evaluations. Such targeted scheduling mitigates the tension between optimization efficiency and morphological diversity inherent in fixed-complexity approaches (see Section \ref{sec:diversity}).

Compared with MorphVAE and LASeR, AdaControl enables a simple GA to rival co-design methods built on sophisticated generative models, demonstrating that principled control scheduling can substitute for search-level complexity. Moreover, the GA-based approach entirely avoids the overhead of training and querying deep generative models, further reducing total computational cost. The contrast with FitControl is equally informative: both methods dynamically adjust control learning, yet AdaControl substantially outperforms FitControl while using considerably fewer PPO iterations (Table \ref{tab:ppo_compare}). We attribute this to a fundamental difference in scheduling philosophy. FitControl adjusts training duration at the individual level based on per-morphology fitness signals, whereas AdaControl operates at the population level, directly targeting the selection bias that arises between individuals. Since fitness evaluation ultimately serves natural selection across candidate solutions, our results suggest that control scheduling guided by population-level indicators is more effective than individual-level heuristics.

\subsubsection{Morphological Diversity}
\label{sec:diversity}
\begin{figure*}[h]
    \centering
    \begin{subfigure}{0.325\textwidth}
        \includegraphics[width=\textwidth]{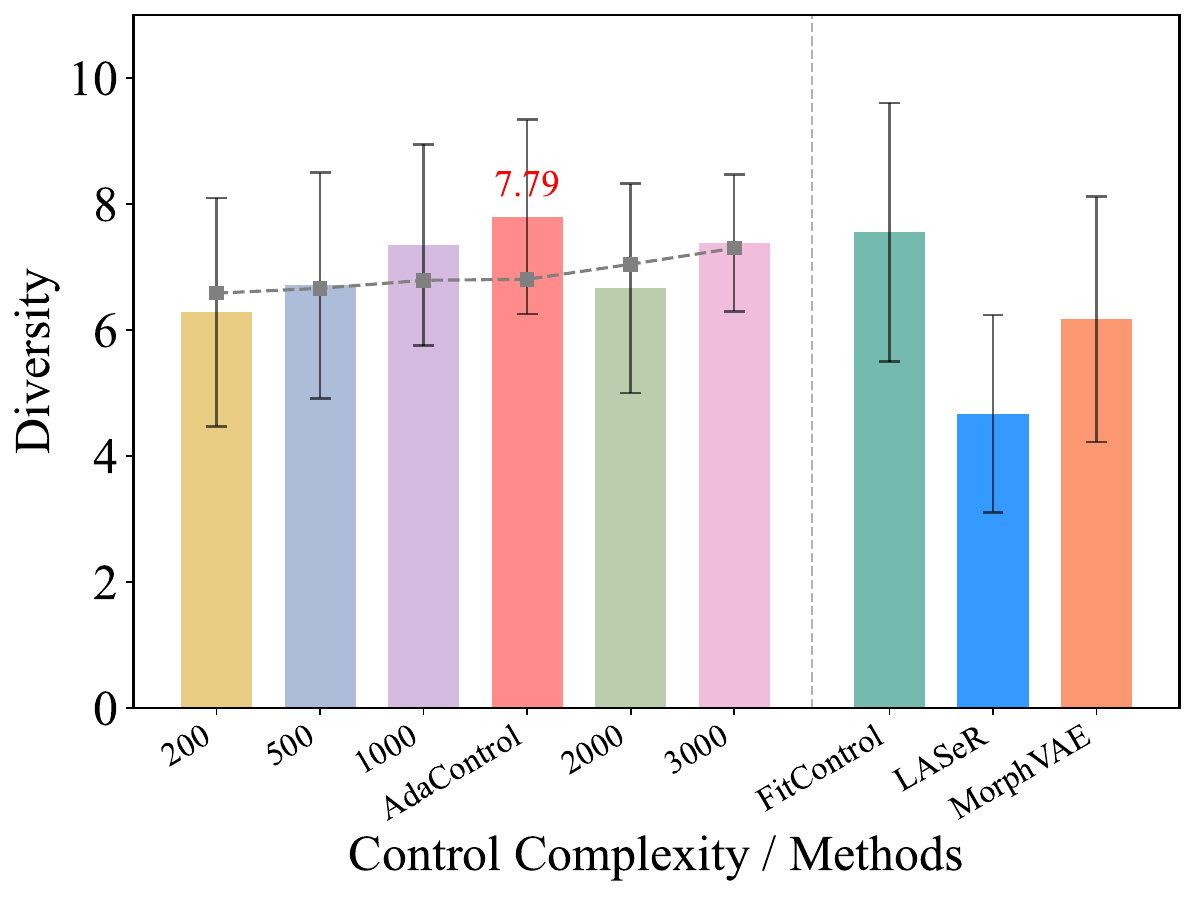}
        \caption{Carrier}
    \end{subfigure}
    \hfill
    \begin{subfigure}{0.32\textwidth}
        \includegraphics[width=\textwidth]{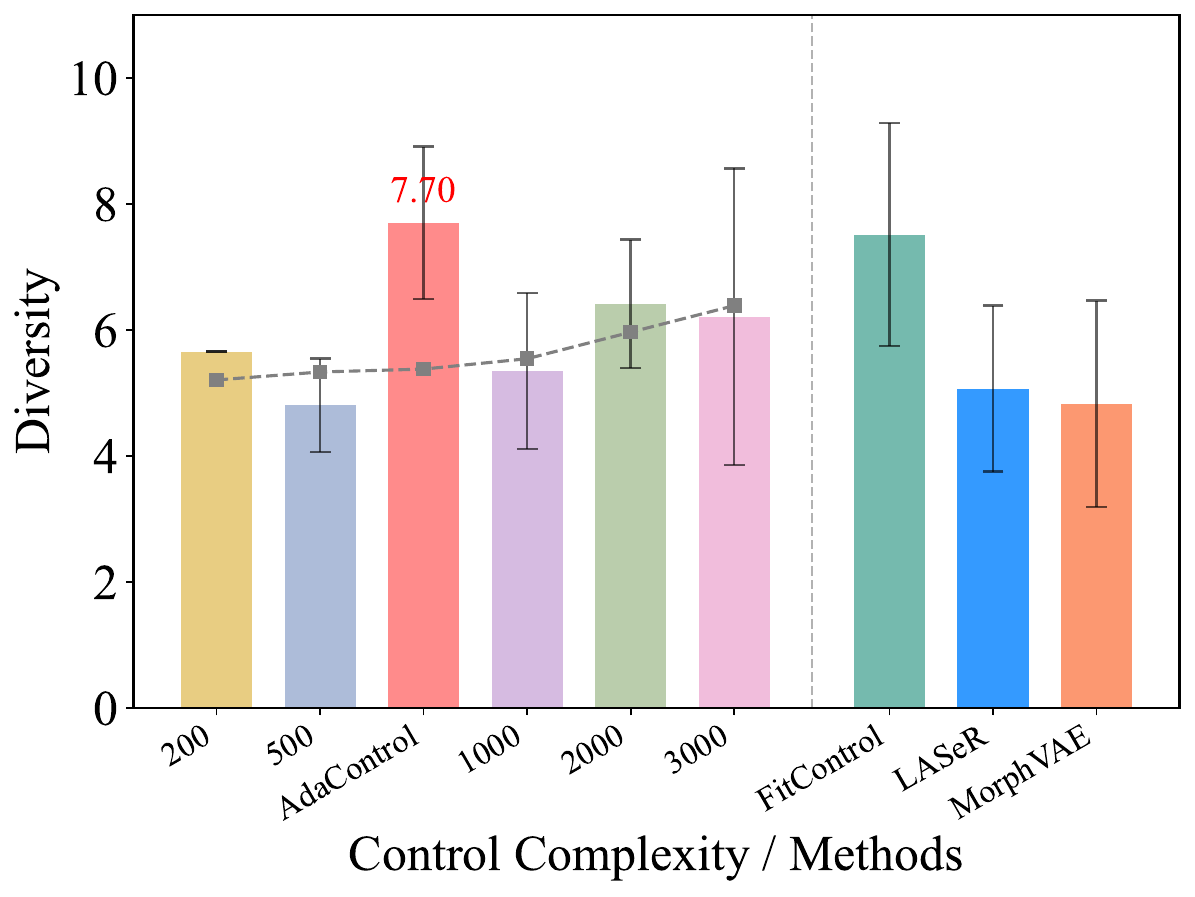}
        \caption{Pusher}
    \end{subfigure}
    \hfill
    \begin{subfigure}{0.32\textwidth}
        \includegraphics[width=\textwidth]{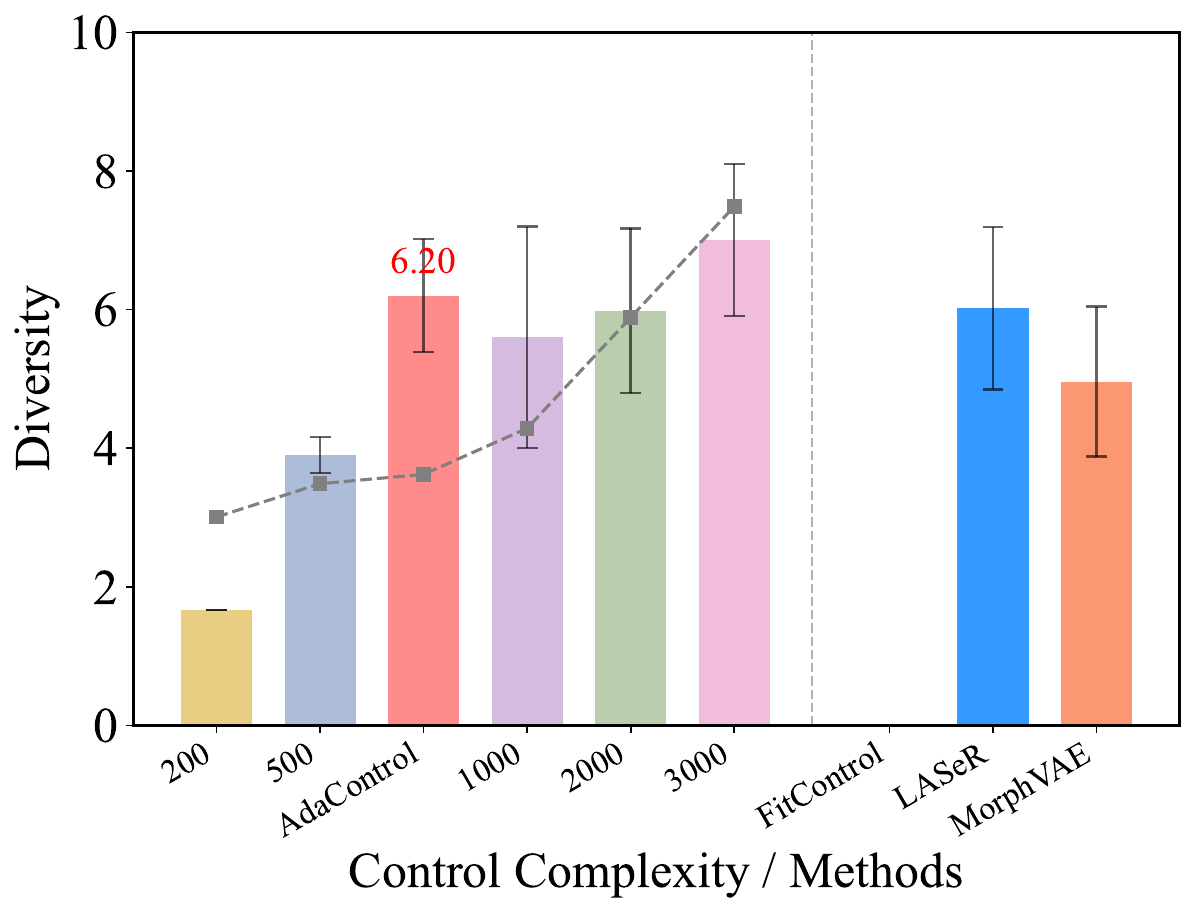}
        \caption{BridgeWalker}
    \end{subfigure}

    \caption{Diversity of high-performing morphologies under different control schemes. Left: GA with fixed complexities and AdaControl, positioned by average PPO iterations, with dashed regression line from fixed-complexity results. Right: FitControl, LASeR, and MorphVAE. FitControl is absent in (c) as it did not produce any high-performing morphology in BridgeWalker.}
    \label{fig:diversity}
\end{figure*}

Morphological diversity offers the most direct window into how evolutionary bias constrains design space exploration. As shown in Fig. \ref{fig:diversity}, diversity of high-performing morphologies increases nearly monotonically with control complexity under fixed schemes, reflecting that more thorough fitness evaluation preserves a wider range of viable evolutionary trajectories. To contextualize AdaControl, we fit linear regression lines to the fixed-complexity results and position AdaControl according to its average PPO iterations. Across all three tasks, AdaControl achieves diversity substantially exceeding the trend predicted by its computational cost. In Carrier and Pusher, AdaControl even surpasses the diversity of strong control while consuming less than half the computation. This diversity gain stems from AdaControl's targeted allocation of learning resources: by investing additional iterations specifically in generations where fast learners dominate selection, it opens evolutionary pathways that uniform training would leave unexplored. As shown in Fig. \ref{fig:diversity}, AdaControl also achieves higher diversity than all baselines, including the state-of-the-art generative-model-based methods MorphVAE and LASeR as well as the adaptive FitControl, corroborating the advantage of population-level control scheduling discussed above.

In summary, conventional co-design with fixed control complexity faces an inherent tension between computational cost and evolutionary performance. AdaControl resolves this by dynamically investing computation where bias is detected, achieving strong optimization efficiency and superior diversity simultaneously. This advantage traces back to the analytical perspective proposed in this work: by extracting morphological properties directly from control learning profiles and tracking their population-level statistics, we uncover the interplay between control learning and selection bias, which in turn naturally motivates AdaControl as a bias-aware scheduling algorithm.

\subsection{Threshold Selection for AdaControl}
\label{sec:threshold}
\begin{figure}[h]
\centering
\includegraphics[width=0.6\textwidth]{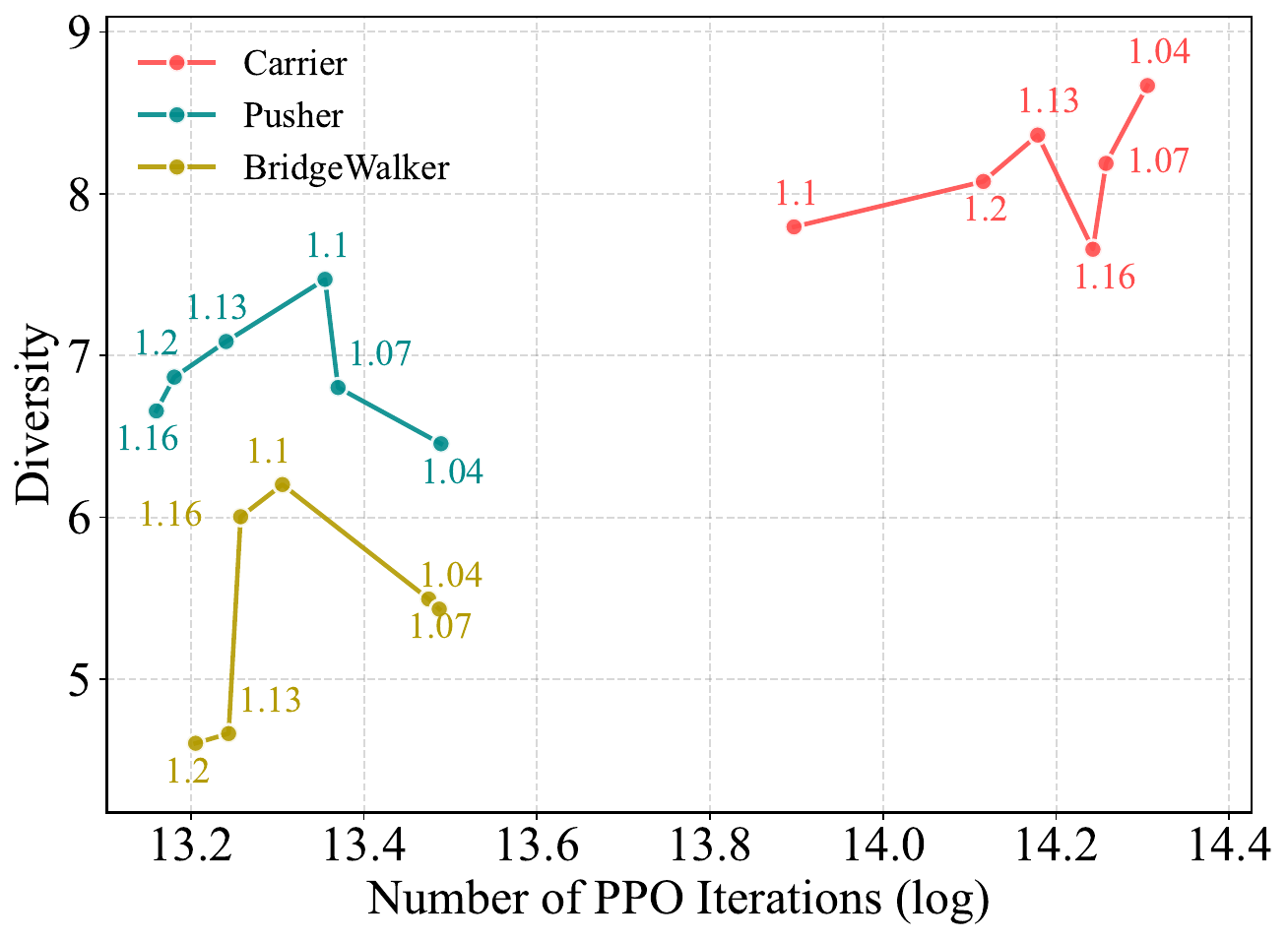}
\caption{Diversity of high-performing morphologies versus total PPO iterations (log scale) for different MI ratio thresholds $r_{\text{thr}}$. Each point is labeled with its threshold value.}
\label{fig:choose_r}
\end{figure}

Rather than setting $r_{\text{thr}}$ subjectively, we adopt a principled selection procedure based on morphological diversity, which directly measures how thoroughly evolution explores the design space. We sweep $r_{\text{thr}} \in \{1.04,\allowbreak 1.07,\allowbreak 1.1,\allowbreak 1.13,\allowbreak 1.16,\allowbreak 1.2\}$ and report diversity against total cumulative PPO iterations in Fig. \ref{fig:choose_r}.

For Pusher and BridgeWalker, diversity peaks at $r_{\text{thr}}=1.1$. Larger thresholds (1.13, 1.16, 1.2) conserve computation but leave MI bias uncorrected, restricting the search to a fast-learner subspace. Contrary to expectation, stricter thresholds (1.07, 1.04) also reduce diversity despite greater computational investment. A plausible explanation is that near-convergent control learning produces highly deterministic fitness rankings, reducing the stochasticity in natural selection that helps sustain population diversity and accelerating convergence along narrow evolutionary paths. This effect appears most pronounced in BridgeWalker, where faster learning convergence (Table \ref{tab:convergence_stats}) makes the population more susceptible to such premature convergence. These results suggest that $r_{\text{thr}}$ shapes evolutionary behaviors in more nuanced ways, simultaneously modulating MI bias and selection stochasticity that affect design space coverage in opposing directions. Diversity peaks where the two are balanced, which also accounts for the residual MI growth in BridgeWalker (Fig. \ref{fig:trace}(c)), where the optimal threshold tolerates moderate bias to preserve selection stochasticity. For Carrier, $r_{\text{thr}}=1.1$ uses the fewest iterations while achieving near-optimal diversity. Other thresholds incur substantially higher computational costs with only marginal diversity changes. Based on these results, $r_{\text{thr}}=1.1$ is selected as the operating point for all experiments.

\subsection{Interpretability Analysis}
Having established the evolutionary bias towards high MI and its implications for co-design, we now conduct a preliminary investigation into the physical underpinnings of morphological intelligence. Taking Carrier as an example, we examine the relationship between MI and three morphological attributes: (a) energy efficiency measured by Cost of Work (COW); (b) the number of empty voxels; (c) the number of soft voxels. Following \citet{gupta2021embodied}, COW is defined as the energy consumed per unit mass to accomplish the task:

\begin{equation}
\label{eq:cow}
\text{COW}=\frac{E}{Mgr},
\end{equation}
where $E$ is the total energy expenditure, measured as the absolute sum of actuation signals; $M$ is the robot mass, measured as the number of non-empty voxels; $r$ is the cumulative episodic reward; and $g$ is the gravitational acceleration, omitted from our calculation as it is constant across all robots.

\begin{figure*}[htb!]
    \centering
    \begin{subfigure}{0.3\textwidth}
        \includegraphics[width=\textwidth]{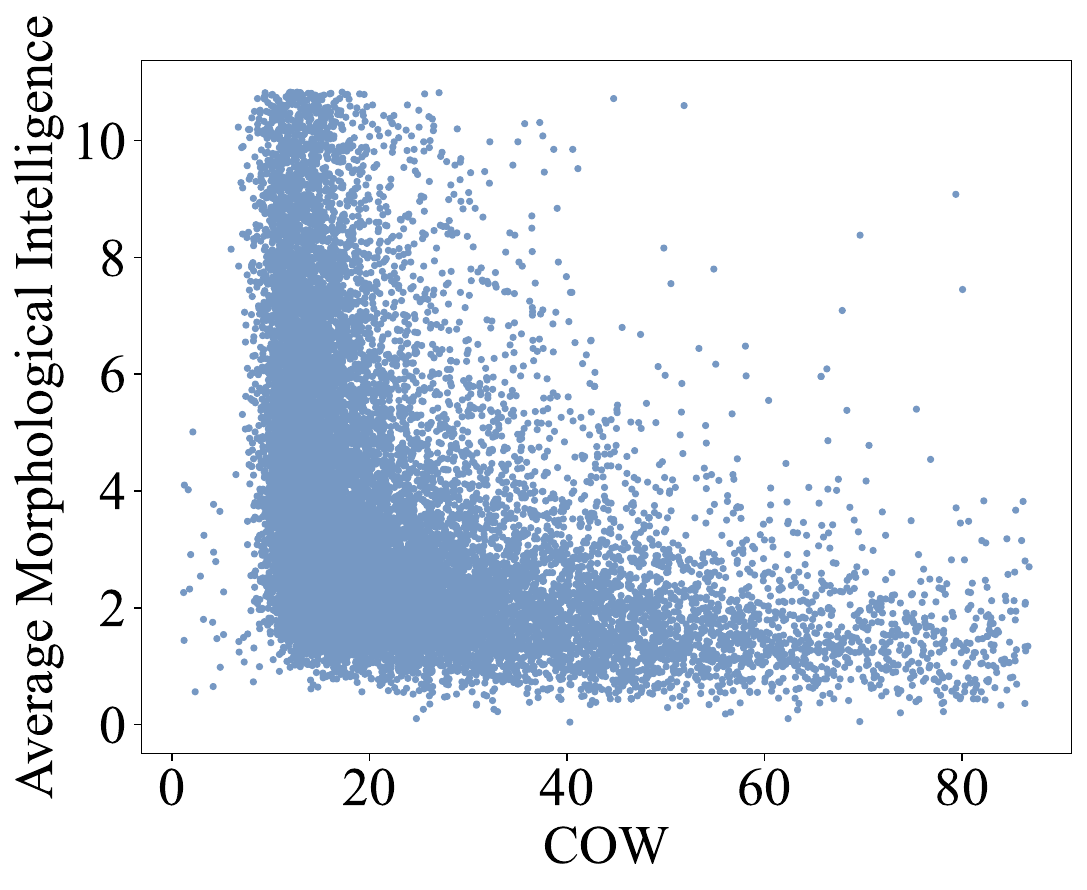}
        \caption{Energy efficiency}
    \end{subfigure}
    \hfill
    \begin{subfigure}{0.3\textwidth}
        \includegraphics[width=\textwidth]{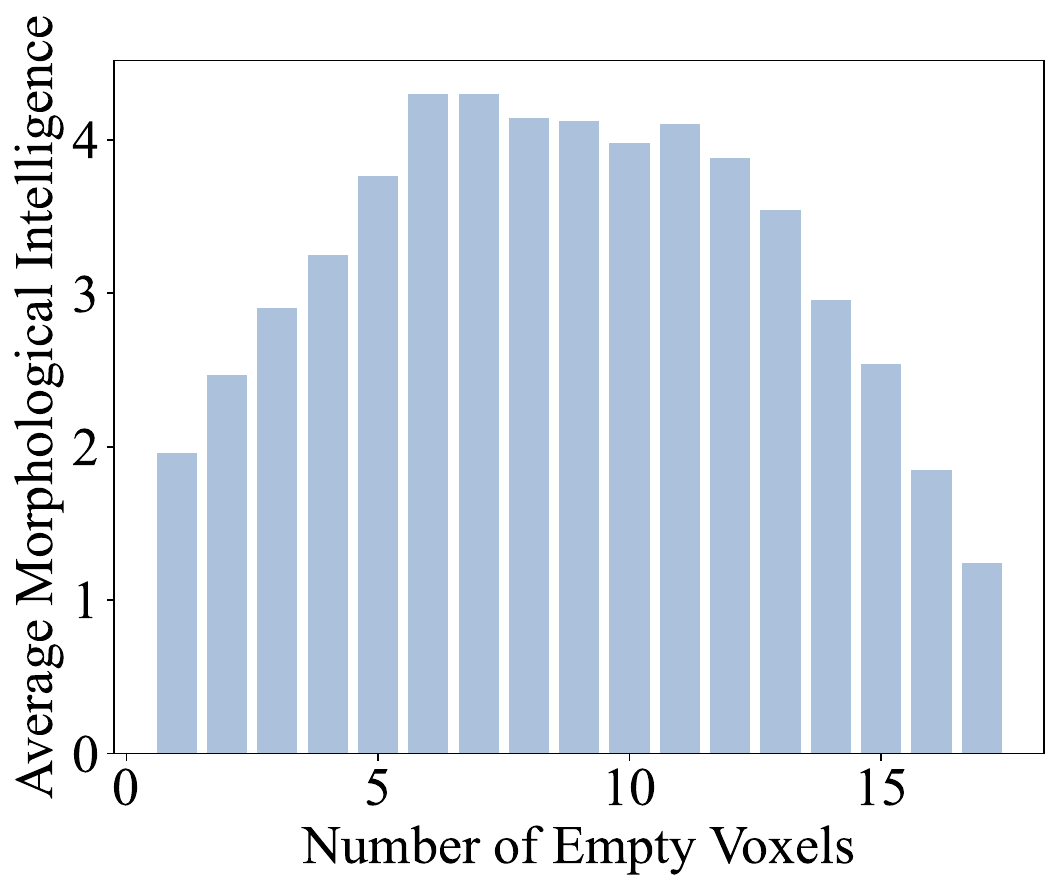}
        \caption{Number of empty voxels}
    \end{subfigure}
    \hfill
    \begin{subfigure}{0.3\textwidth}
        \includegraphics[width=\textwidth]{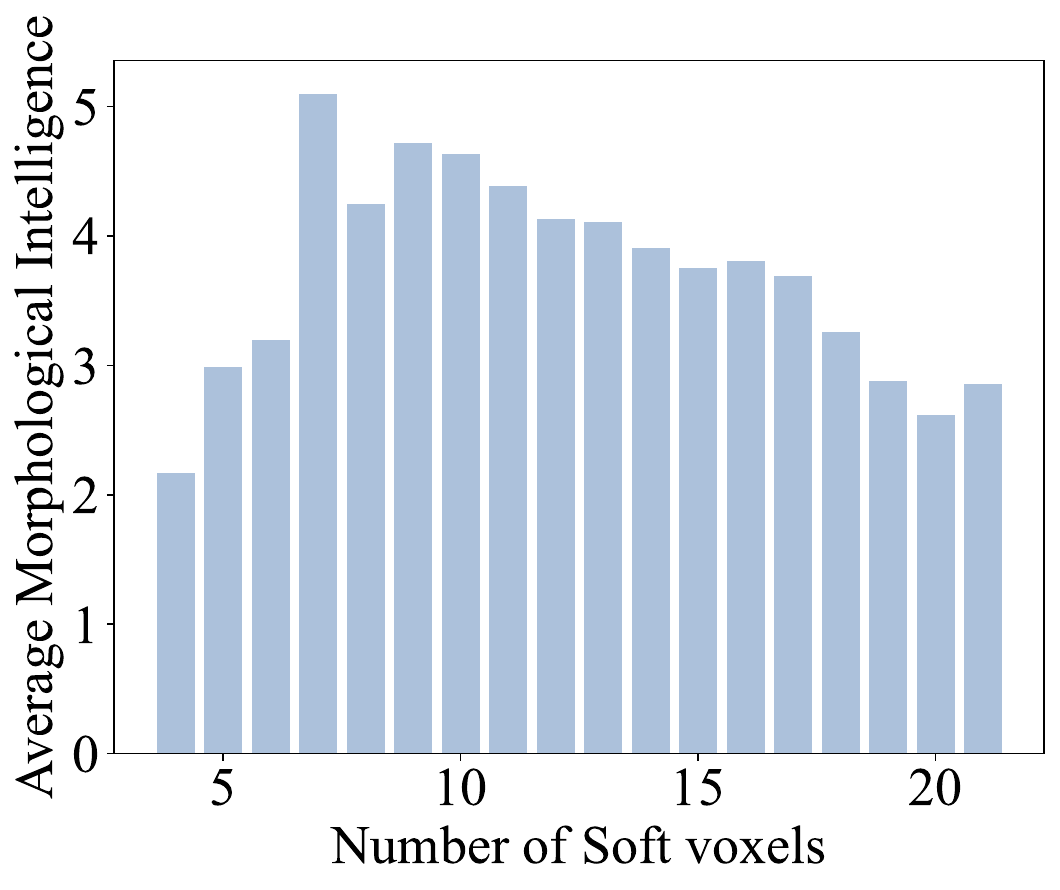}
        \caption{Number of soft voxels}
    \end{subfigure}

    \caption{Relationship between MI and morphological attributes in Carrier. (a) Each point represents one morphology. (b)--(c) Morphologies with the same voxel count are aggregated; only the mean MI is shown.}
    \label{fig:interpretable}
\end{figure*}

As shown in Fig. \ref{fig:interpretable}(a), robots with higher MI exhibit lower COW, mirroring the pattern observed for rigid robots in \citet{gupta2021embodied} and suggesting that morphologically intelligent soft robots are better able to exploit passive body-environment dynamics for energy-efficient behavior. The relationship between MI and the number of empty voxels follows an inverted U-shape (Fig. \ref{fig:interpretable}(b)). A moderate number of empty voxels appears to reduce structural constraints and enable more compliant deformations, facilitating easier control. Beyond a certain point, however, overly sparse structures may give rise to interaction dynamics too complex to be effectively exploited. A similar non-monotonic pattern is observed for soft voxels (Fig. \ref{fig:interpretable}(c)), partly consistent with \citet{corucci2016material}, with the decline at higher counts admitting a similar explanation.

These findings suggest that energy efficiency and structural composition are key physical attributes underpinning MI in voxel-based soft robots, and offer concrete insight into which regions of the design space are favored by biased evolutionary processes.

\section{Conclusion}
\label{sec:conclusion}
In this work, we investigate the brain-body co-evolution of learning-based robotic systems across two timescales, and reveal that the configuration of control learning is a critical yet overlooked determinant of morphological evolution. By decomposing the intrinsic learning profile of morphologies into morphological intelligence and true potential, we provide a quantitative framework that exposes how prematurely terminated control learning biases selection and gives rise to the morphological Baldwin effect as a special case. AdaControl, grounded in population-level MI monitoring, resolves this bias with minimally sufficient computation and demonstrates that evaluation fidelity, rather than search sophistication, is the primary bottleneck in co-design. Our threshold selection analysis further uncovers a dual role of the MI ratio threshold in governing both MI bias and selection stochasticity, offering a nuanced understanding of how control scheduling shapes evolutionary exploration.

Our findings are established on simulated voxel-based soft robots across three tasks. Whether our findings generalize to other morphological representations, task domains, and physical platforms remains to be verified \citep{wang2025codesign_survey,stolzle2025holistic}. On the algorithmic side, our threshold selection analysis reveals task-dependent behavior, motivating the development of adaptive threshold mechanisms that self-calibrate during evolution. More broadly, control complexity extends beyond training duration to encompass network architecture and learning algorithms, each of which may interact with morphological evolution in distinct ways that our framework is well positioned to investigate further.

\section*{Acknowledgments}
The authors would like to thank Prof.\ Feifei Wang for her valuable guidance and Zhongmin Liang for her contributions to this work. This work is supported by the Intelligent Game and Decision Laboratory and the Zhiqiang Foundation.

\end{document}